\documentclass[journal]{IEEEtran}

\usepackage{blindtext}

\usepackage{amsmath,amssymb,amsfonts,bm}
\usepackage{lipsum}
\usepackage{times}
\usepackage{epsfig}

\usepackage{stfloats}
\usepackage{multicol}
\usepackage[inline]{enumitem}
\usepackage{algorithmic}
\usepackage{graphicx,subcaption}
\usepackage{textcomp}

\usepackage{booktabs}
\usepackage{multirow}

\usepackage[utf8]{inputenc}
\usepackage[T1]{fontenc}
\usepackage{float}
\usepackage{adjustbox}
\usepackage{array}
\usepackage{dsfont}

\usepackage{paralist}
\usepackage[ruled,linesnumbered]{algorithm2e}
\usepackage[accsupp]{axessibility}  

\usepackage{tabulary}
\usepackage[table]{xcolor}

\definecolor{gray}{rgb}{0.3,0.3,0.3}
\definecolor{blue}{rgb}{0,0.5,1}
\definecolor{mask_red}{rgb}{1,0,0.8}
\definecolor{green}{rgb}{0.2,1,0.2}
\definecolor{rblue}{rgb}{0,0,1}

\usepackage{hyperref}
\hypersetup{colorlinks=true,linkcolor=red,citecolor=blue}

\makeatletter
\DeclareRobustCommand\onedot{\futurelet\@let@token\@onedot}
\def\@onedot{\ifx\@let@token.\else.\null\fi\xspace}

\def\eg{\emph{e.g}\onedot} 
\def\ie{\emph{i.e}\onedot} 
 
\def\etc{\emph{etc}\onedot} 
 
\def\etal{\emph{et al}\onedot}

\begin{document}

\title{CMX: Cross-Modal Fusion for RGB-X Semantic Segmentation with Transformers}
\author{Jiaming Zhang\IEEEauthorrefmark{1}, Huayao Liu\IEEEauthorrefmark{1}, Kailun Yang\IEEEauthorrefmark{1}\IEEEauthorrefmark{2}, Xinxin Hu, Ruiping Liu, and Rainer Stiefelhagen
\IEEEcompsocitemizethanks{
\IEEEcompsocthanksitem 
This work was supported in part by the Federal Ministry of Labor and Social Affairs (BMAS) through the AccessibleMaps project under Grant 01KM151112, in part by the ``KIT Future Fields'' project, in part by the MWK through the Cooperative Graduate School Accessibility through AI-based Assistive Technology (KATE) under Grant BW6-03, and in part by the BMBF through a fellowship within the IFI program of the German Academic Exchange Service (DAAD), in part by the HoreKA@KIT supercomputer partition, and in part by Hangzhou SurImage Technology Company Ltd.
\IEEEcompsocthanksitem J. Zhang, R. Liu, and R. Stiefelhagen are with Karlsruhe Institute of Technology, 76131 Karlsruhe, Germany.
\IEEEcompsocthanksitem K. Yang is with Hunan University, Changsha 410082, China.
\IEEEcompsocthanksitem 
H. Liu is with NIO, Shanghai 201804, China.
\IEEEcompsocthanksitem 
X. Hu is with ByteDance Inc., Hangzhou 310000, China.
\IEEEcompsocthanksitem \IEEEauthorrefmark{1}indicates equal contribution.
\IEEEcompsocthanksitem \IEEEauthorrefmark{2}corresponding author. (E-Mail: kailun.yang@hnu.edu.cn.)
}
}

\maketitle
\bstctlcite{IEEEexample:BSTcontrol}
%
\begin{abstract}
Scene understanding based on image segmentation is a crucial component of autonomous vehicles. Pixel-wise semantic segmentation of RGB images can be advanced by exploiting complementary features from the supplementary modality (\emph{X}-modality). However, covering a wide variety of sensors with a modality-agnostic model remains an unresolved problem due to variations in sensor characteristics among different modalities. 
Unlike previous modality-specific methods, in this work, we propose a unified fusion framework, \emph{CMX}, for RGB-X semantic segmentation.
To generalize well across different modalities, that often include supplements as well as uncertainties, a unified cross-modal interaction is crucial for modality fusion. Specifically, we design a Cross-Modal Feature Rectification Module (\emph{CM-FRM}) to calibrate bi-modal features by leveraging the features from one modality to rectify the features of the other modality.
With rectified feature pairs, we deploy a Feature Fusion Module (\emph{FFM}) to perform sufficient exchange of long-range contexts before mixing.
To verify CMX, for the first time, we unify five modalities complementary to RGB, \ie, depth, thermal, polarization, event, and LiDAR. 
Extensive experiments show that CMX generalizes well to diverse multi-modal fusion, 
achieving state-of-the-art performances on five RGB-Depth benchmarks, as well as RGB-Thermal, RGB-Polarization, and RGB-LiDAR datasets. Besides, to investigate the generalizability to dense-sparse data fusion, we establish an RGB-Event semantic segmentation benchmark based on the EventScape dataset, on which CMX sets the new state-of-the-art. The source code of CMX is publicly available at \url{https://github.com/huaaaliu/RGBX_Semantic_Segmentation}.

\end{abstract}


\IEEEpeerreviewmaketitle
\section{Introduction}
\IEEEPARstart{S}{cene} understanding is a fundamental component in Autonomous Vehicles (AVs) since it can provide comprehensive information to support the Advanced Driver-Assistance System (ADAS) to make correct decisions when interacting with the driving surrounding~\cite{zhou2019automated}. As exteroceptive sensors, cameras are adopted in AVs for perceiving the surroundings~\cite{yang2020omnisupervised}.
Image semantic segmentation -- a fundamental task in computer vision -- is an ideal perception solution to transform an image input into its underlying semantically meaningful regions, providing pixel-wise dense scene understanding for Intelligent Transportation Systems (ITS)~\cite{sun2020rfnet,zhang2022trans4trans_tits}.
Image semantic segmentation has made significant progress on accuracy~\cite{chen2017deeplab,zhao2017pspnet,fu2019danet}.
Yet, current models may struggle to extract high-quality features in certain circumstances, \eg, when two objects have similar colors or textures, leading to difficulty in distinguishing them through pure RGB images~\cite{hu2019acnet}.

\begin{figure}[t]
\begin{center}
    \includegraphics[width=0.9\linewidth]{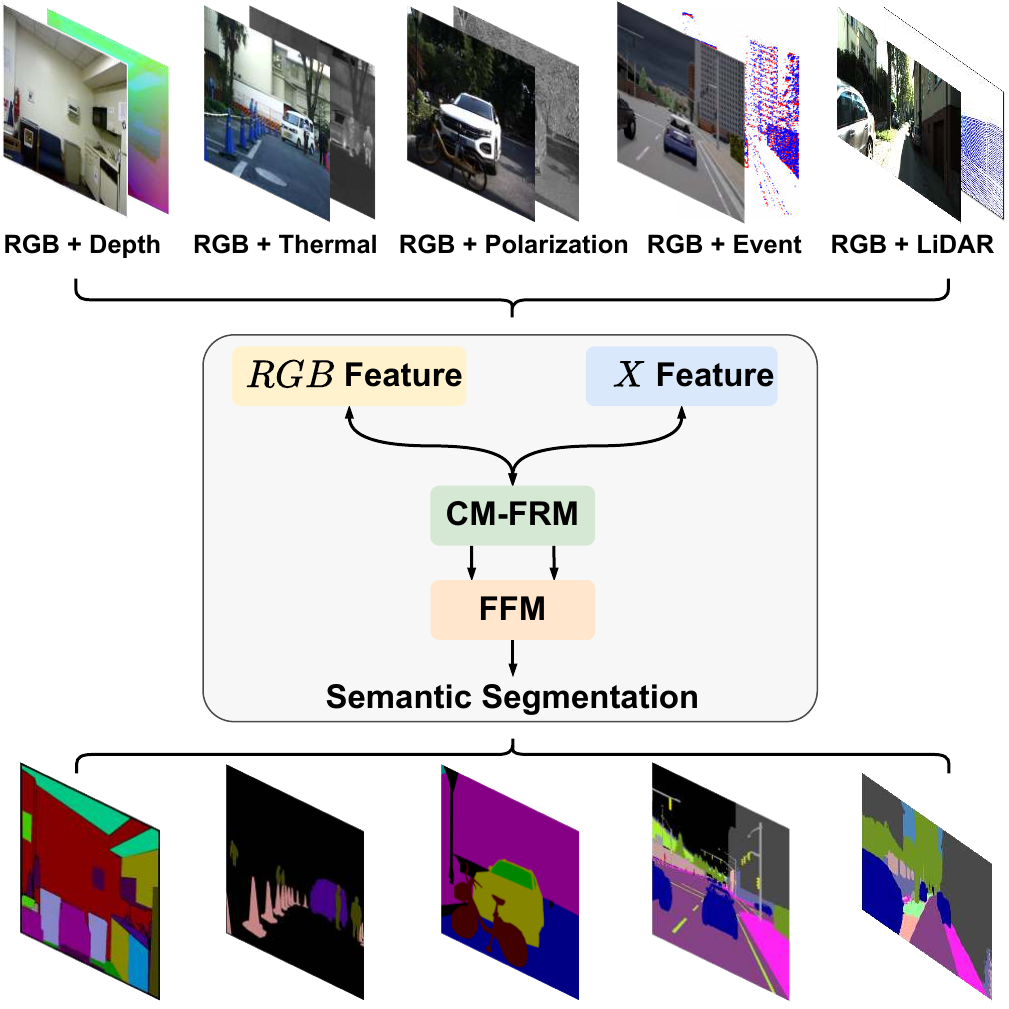}
\end{center}
    \caption{RGB-X semantic segmentation unifies diverse sensing modality combinations: RGB-Depth, -Thermal, -Polarization, -Event, and -LiDAR segmentation. 
    CMX is established with Cross-Modal Feature Rectification Module (\emph{CM-FRM}) to calibrate the features of RGB- and X-modality and Feature Fusion Module (\emph{FFM}) to perform the exchange of long-range context and combine features for RGB-X semantic segmentation.}
\label{fig:intro}
\end{figure}

Thanks to the development of sensor technologies, there is a growing variety of modular sensors which are highly applicable for ITS applications.
Different types of sensors can supply RGB images with rich complementary information (see Fig.~\ref{fig:intro}).
For example, \emph{depth} measurement can help identify the boundaries of objects and offer geometric information of dense scene elements~\cite{hu2019acnet,chen2020sa_gate}.
\emph{Thermal} images facilitate to discern different objects through their specific infrared imaging~\cite{ha2017mfnet,zhang2021abmdrnet}.
Besides, \emph{polarimetric}- and \emph{event} information are advantageous for perception in specular- and dynamic real-world scenes~\cite{xiang2021polarization,zhang2021issafe}. \emph{LiDAR} data can provide spatial information in driving scenarios~\cite{zhuang2021pmf}. Thereby, a research question arises: {\textit{How to construct a unified model to incorporate the fusion of RGB with various modalities, \ie, RGB-X semantic segmentation as illustrated in Fig.~\ref{fig:intro}?}}

\begin{figure}[t]
    \centering
    \includegraphics[width=0.99\columnwidth]{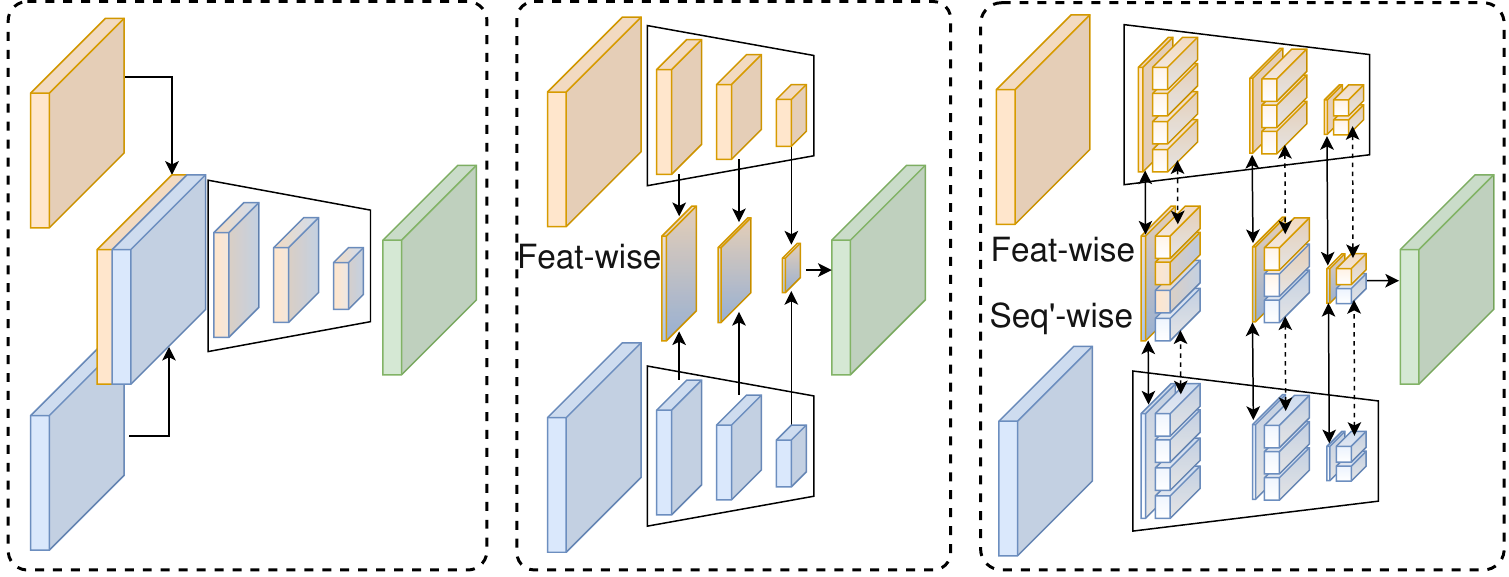}
    \begin{minipage}[t]{.3\columnwidth}\vskip -3ex
        \subcaption{Input fusion}\label{fig2_1:input_fusion}
    \end{minipage}%
    \begin{minipage}[t]{.32\columnwidth}\vskip -3ex
        \subcaption{Feature fusion}\label{fig2_2:feat_fusion}
    \end{minipage}%
    \begin{minipage}[t]{.34\columnwidth}\vskip -3ex
        \subcaption{Interactive fusion}\label{fig2_3:interact_fusion}
    \end{minipage}%
    \caption{Comparison of different fusion methods. (a) Input fusion merges inputs with modality-specific operations~\cite{cao2021shapeconv,chen2021spatial_guided}. {(b) Feature fusion applies channel attention to fuse features in a unidirectional manner~\cite{hu2019acnet,chen2020sa_gate}.
    (c) Our interactive fusion incorporates bidirectional cross-modal feature rectification, and sequence-to-sequence cross-attention, yielding comprehensive cross-modal interactions.}
   }
\label{fig:comparison_fusion_methods}
\end{figure}

\begin{figure}[t]
    \centering
   \includegraphics[width=1.0\linewidth, keepaspectratio]
   {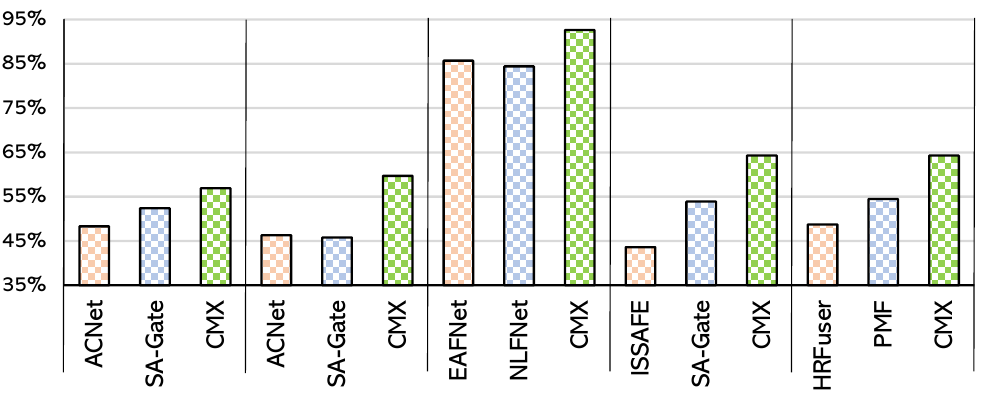}
    \begin{minipage}[t]{.2\columnwidth}
        \vskip -3ex
        \subcaption{RGB-D}\label{fig3_1:rgbd}
    \end{minipage}%
    \begin{minipage}[t]{.2\columnwidth}
        \vskip -3ex
        \subcaption{RGB-T}\label{fig3_2:rgbt}
    \end{minipage}%
    \begin{minipage}[t]{.2\columnwidth}
        \vskip -3ex
        \subcaption{RGB-P}\label{fig3_3:rgbp}
    \end{minipage}%
    \begin{minipage}[t]{.2\columnwidth}
        \vskip -3ex
        \subcaption{RGB-E}\label{fig3_4:rgbe}
    \end{minipage}%
    \begin{minipage}[t]{.2\columnwidth}
        \vskip -3ex
        \subcaption{RGB-L}\label{fig3_5:rgbl}
    \end{minipage}%
   \caption{Performance comparison on different RGB-X semantic segmentation benchmarks. SA-Gate~\cite{chen2020sa_gate} designed for RGB-D data (\eg, on NYU Depth V2 dataset~\cite{silberman2012nyu_dataset}), is less effective on RGB-T or RGB-E tasks. Our modality-agnostic CMX, for the first time, outperforms modality-specific methods on five segmentation tasks.
   } 
    \label{fig:performance_comparison}
\end{figure}

Existing multi-modal semantic segmentation methods can be divided into two categories: (1) The first category~\cite{cao2021shapeconv,chen2021spatial_guided} employs a single network to extract features from RGB and another modality, which are fused in the input stage (see Fig.~\ref{fig2_1:input_fusion}).
(2) The second type of approaches~\cite{chen2020sa_gate,zhang2021abmdrnet,deng2021feanet} deploys two backbones to perform feature extraction from RGB- and another modality separately then fuses the extracted two features into one feature for semantic prediction (see Fig.~\ref{fig2_2:feat_fusion}).
{However, both types are usually well-tailored for a single specific modality pair (\eg, RGB-D or RGB-T), yet hard to be extended to operate with other modality combinations. For example, regarding our observation in Fig.~\ref{fig:performance_comparison}, ACNet~\cite{hu2019acnet} and SA-Gate~\cite{chen2020sa_gate}, designed for RGB-D data, perform less satisfactorily in RGB-T tasks.}
{To flexibly cover various sensor combinations for ITS applications, a unified \emph{RGB-X semantic segmentation}, is desirable and advantageous.} {Its benefits are two-fold:
(1) It can save research and engineering efforts, with no need to adapt architectures for a specific modality combination scenario.
(2) It makes it possible that a system equipped with multi-modal sensors can readily leverage new sensors when they become available~\cite{sun2019multimodal,girdhar2022omnivore}, which is conducive to robust scene perception. For this purpose, in this work, we spend efforts to construct a modality-agnostic framework for unified RGB-X semantic segmentation.}

Recently, vision transformers~\cite{vaswani2017attention,dosovitskiy2021vit,touvron2021deit,liu2021swin} handle inputs as sequences and are able to acquire long-range correlations, offering the possibility for a unified framework for diverse multi-modal tasks. Compared to existing multi-modal fusion modules~\cite{hu2019acnet,xiang2021polarization,deng2021feanet} based on Convolutional Neural Networks (CNNs), it remains unclear whether potential improvements on RGB-X semantic segmentation can be materialized via vision transformers. 
Crucially, while some previous works~\cite{hu2019acnet,chen2020sa_gate} use a simple global multi-modal interaction strategy, it does not generalize well across different sensing data combinations~\cite{zhang2021abmdrnet}. We hypothesize that for RGB-X semantic segmentation with various supplements and uncertainties, comprehensive cross-modal interactions should be provided, to fully exploit the potential of cross-modal complementary features.

To tackle the aforementioned challenges, 
we propose \emph{CMX}, a universal cross-modal fusion framework for RGB-X semantic segmentation in an interactive fusion manner (Fig.~\ref{fig2_3:interact_fusion}).
Specifically, CMX is built as a two-stream architecture, \ie, RGB- and X-modal stream.
Two specific modules are designed for feature interaction and feature fusion in between.
{(1) \emph{Cross-Modal Feature Rectification Module (CM-FRM)}, calibrates the bi-modal features by leveraging their spatial- and channel-wise correlations, which enables both streams to focus more on the complementary informative cues from each other, as well as mitigates the effects of uncertainties and noisy measurements from different modalities.
Such a feature rectification tackles varying noises and uncertainties in diverse modalities. It enables better multi-modal feature extraction and interaction.
(2) \emph{Feature Fusion Module (FFM)}, {is constructed in two stages and it performs sufficient information exchange before merging features.}
Motivated by the large receptive fields obtained via self-attention~\cite{vaswani2017attention}, a cross-attention mechanism is devised in the first stage of FFM for realizing cross-modal global reasoning.
In its second stage, mixed channel embedding is applied to produce enhanced output features.
Thereby, our introduced comprehensive interactions lie in multiple levels (see Fig.~\ref{fig2_3:interact_fusion}). It includes channel- and spatial-wise rectification from the feature map perspective, as well as cross-attention from the sequence-to-sequence perspective, which are critical for generalization across modality combinations.}

To verify our unification proposal, we consider and assess CMX on $5$ different multi-modal semantic segmentation tasks, including RGB-Depth, -Thermal, -Polarization, -Event, and -LiDAR semantic segmentation. A total of $9$ datasets are involved.
In particular, CMX attains top mIoU of $56.9\%$ on NYU Depth V2 (RGB-D)~\cite{silberman2012nyu_dataset}, $59.7\%$ on MFNet (RGB-T)~\cite{ha2017mfnet}, $92.6\%$ on ZJU-RGB-P (RGB-P)~\cite{xiang2021polarization}, and $64.3\%$ on KITTI-360 (RGB-L)~\cite{liao2021kitti360} datasets.
Our universal approach CMX clearly outperforms specialized architectures (Fig.~\ref{fig:performance_comparison}).
Furthermore, to address the lack of RGB-Event parsing benchmark in the community, 
we establish an RGB-Event semantic segmentation benchmark based on the EventScape dataset~\cite{gehrig2021eventscape_dataset}, where our CMX sets the new state-of-the-art among ${>}10$ benchmarked models.
Besides, our experiments demonstrate that the CMX framework is effective for both CNN- and Transformer-based architectures.
Moreover, our investigation on representations of polarization- and event-based data indicates the path to follow and the sweet spot for reaching robust multi-modal semantic segmentation, trumping original representation methods~\cite{xiang2021polarization,gehrig2021eventscape_dataset}.

At a glance, we deliver the following contributions:
\begin{compactitem}
    \item {For the first time, we explore \emph{RGB-X semantic segmentation} in five types of multi-modal sensing data combinations, including RGB-Depth, RGB-Thermal, RGB-Polarization, RGB-Event, and RGB-LiDAR.}
    \item {We rethink multi-modality fusion from a generalization perspective and prove that comprehensive cross-modal interaction is crucial for the unification of fusion across diverse modalities.}
    \item We propose an RGB-X semantic segmentation framework \emph{CMX} with \emph{cross-modal feature rectification} and \emph{feature fusion} modules, intertwining cross-attention and mixed channel embedding for enhanced global reasoning.
    \item {We investigate different representations of polarimetric- and event data and indicate the optimal path to follow for reaching robust multi-modal semantic segmentation.}
    \item {An RGB-Event semantic segmentation benchmark is established to assess dense-sparse data fusion, and is incorporated into the RGB-X semantic segmentation.} 
\end{compactitem}

\section{Related Work}
\subsection{Transformer-driven Semantic Segmentation}
For dense semantic segmentation, pyramid-, strip-, and atrous spatial pyramid pooling are designed to harvest multi-scale feature representations~\cite{chen2017deeplab,zhao2017pspnet}. 
Besides, cross-image pixel contrast learning~\cite{wang2021exploring_contrast} is applied to address intra-class compactness and inter-class dispersion, while nonparametric nearest prototype retrieving~\cite{zhou2022rethinking} is proposed to achieve semantic segmentation in a prototype view.
Inspired by the non-local block~\cite{wang2018nonlocal}, self-attention in transformers~\cite{vaswani2017attention} has been used to establish long-range dependencies by DANet~\cite{fu2019danet}
and CCNet~\cite{huang2019ccnet}.
Recently, SETR~\cite{zheng2021setr} and Segmenter~\cite{strudel2021segmenter} directly adopt vision transformers~\cite{dosovitskiy2021vit,touvron2021deit} as the backbone, which {captures} global context from very early layers. 
SegFormer~\cite{xie2021segformer} and Swin~\cite{liu2021swin} create hierarchical structures to make use of multi-resolution features.
Following this trend, various architectures of dense prediction transformers~\cite{wang2021pvt,yuan2021hrformer} and semantic segmentation transformers~\cite{zhang2022early_region_proxy,lin2022structtoken} emerge in the field.
While these approaches have achieved high performance, most of them focus on using RGB images and suffer when RGB images cannot provide sufficient information in real-world scenes, \eg, under low-illumination conditions or in high-dynamic areas. 
In this work, we tackle multi-modal semantic segmentation to take advantage of complementary information from other modalities such as depth, thermal, polarization, event, and LiDAR data for boosting RGB segmentation.

\subsection{Multi-modal Semantic Segmentation}
While previous works reach high performance on standard RGB-based semantic segmentation benchmarks, in challenging real-world conditions, it is desirable to involve multi-modality sensing for a reliable and comprehensive scene understanding.
{RGB-Depth~\cite{qian2021rfbnet_gated,zhou2021canet} and RGB-Thermal~\cite{sun2019rtfnet,sun2020fuseseg,zhou2021gmnet} semantic segmentation are broadly investigated.
Polarimetric optical cues~\cite{kalra2020deep_polarization} and event-driven priors~\cite{zhang2021edcnet} are often intertwined for robust perception under adverse conditions.
In automated driving, LiDAR data~\cite{zhuang2021pmf} is incorporated for enhanced semantic road scene understanding.
However, most of these works only address a single modality combination.
In this work, we explore a unified approach, which can generalize well to diverse multi-modal combinations.}

For multi-modal semantic segmentation, there are two dominant strategies. The first mainstream paradigm models cross-modal complementary information into layer- or operator designs~\cite{cao2021shapeconv,chen2021spatial_guided,wang2018depth_aware,xing2020malleable,wu2020depth_adapted}.
While these works verify that multi-modal features can be learned within a shared network, they are carefully designed for a single modality, \eg, RGB-D semantic segmentation, which is hard to be applied to other modalities.
Moreover, there are multi-task frameworks~\cite{zhang2019pattern,bachmann2022multimae} that facilitate inter-task feature propagation for RGB-D scene understanding, but they rely on supervision from other tasks for joint learning.
The second paradigm dedicates to developing fusion schemes to bridge two parallel modality streams.
ACNet~\cite{hu2019acnet} proposes attention modules to exploit informative features for RGB-D semantic segmentation, whereas ABMDRNet~\cite{zhang2021abmdrnet} suggests reducing the modality differences of features before selectively extracting discriminative cues for RGB-T fusion. 
For RGB-P segmentation, Xiang~\etal~\cite{xiang2021polarization} connect RGB- and polarization branches via channel attention bridges.
For RGB-E parsing, Zhang~\etal~\cite{zhang2021issafe} explore sparse-to-dense and dense-to-sparse fusion flows to extract dynamic context for accident scene segmentation.
Salient object detection, seen as a specific type of image segmentation, can also benefit from multimodal fusion to identify the most important objects, such as Hyperfusion-Net~\cite{zhang2019hyperfusion} tailored for RGB-D and CAVER~\cite{pang2023caver} for RGB-D and RGB-T.
In this research, we also advocate this paradigm but unlike previous works, we address RGB-X semantic segmentation with a unified framework, for generalizing to diverse sensing modality combinations.

While previous works use a simple global channel-wise strategy, it does not work well across different sensing data. For example, ACNet~\cite{hu2019acnet} and SA-Gate~\cite{chen2020sa_gate}, designed for RGB-D segmentation, perform less satisfactorily in RGB-T scene parsing~\cite{zhang2021abmdrnet}. In contrast, we hypothesize that comprehensive cross-modal interactions are crucial for RGB-X semantic segmentation with various supplements and uncertainties, so as to fully unleash the potential of cross-modal complementary features. Besides, most of the previous works adopt CNN backbone without considering that long-range dependency.
We put forward a framework with transformers, which has global dependencies already in its architecture design.
Differing from existing works, we perform fusion on different levels with cross-modal feature rectification and cross-attentional exchanging for enhanced dense semantic prediction.

\begin{figure*}[t]
\begin{center}
    \includegraphics[width=1\linewidth]{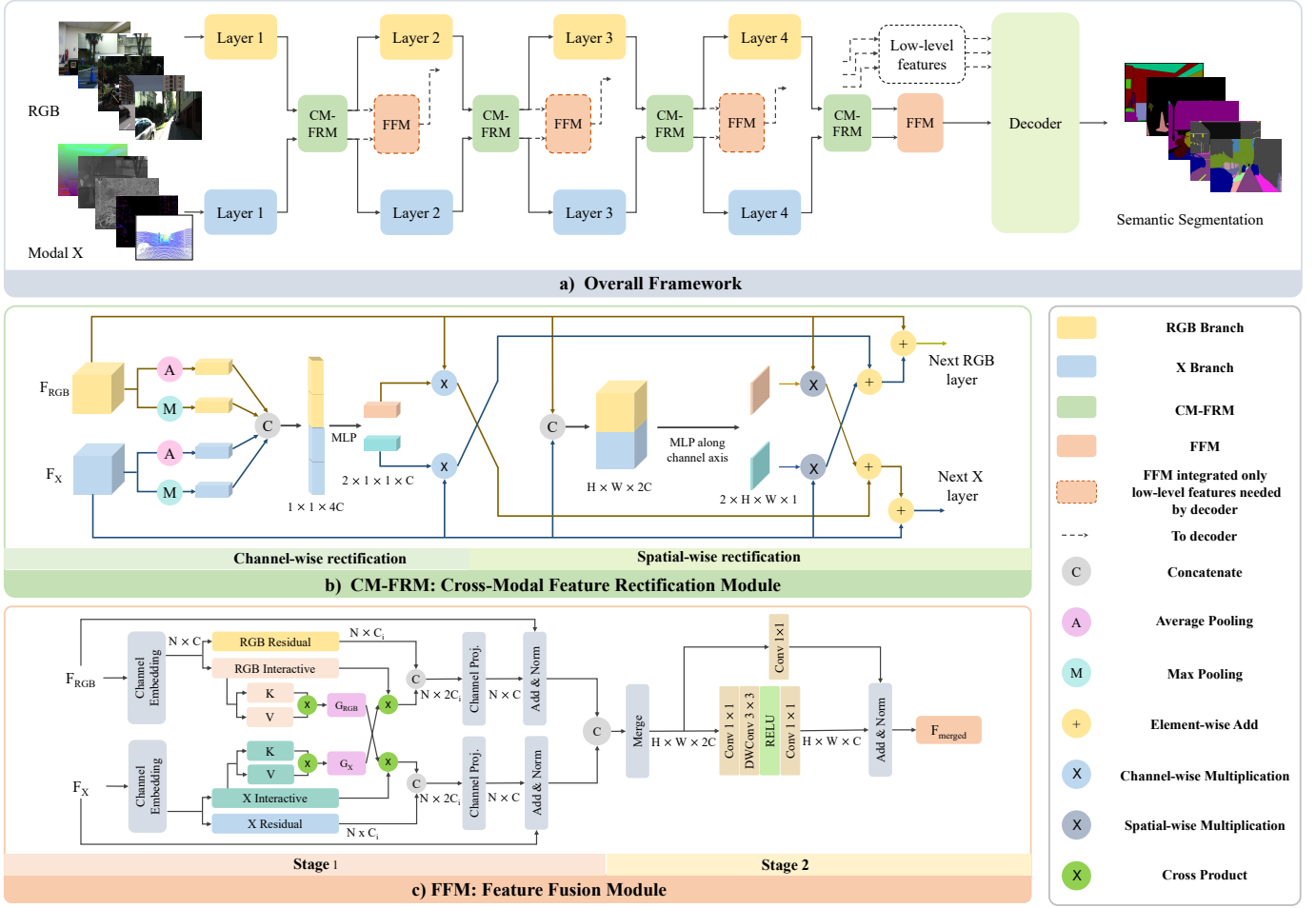}
\end{center}
   \caption{{a) Overview of \emph{CMX} for \emph{RGB-X semantic segmentation}. The inputs are RGB and another modality (\eg, Depth, Thermal, Polarization, Event, or LiDAR). b) Cross-Modal Feature Rectification Module (\emph{CM-FRM}) with colored arrows as information flows of the two modalities. c) Feature Fusion Module (\emph{FFM}) with two stages of information exchange and fusion.}
   }
\label{fig:framework}
\end{figure*}

\section{Proposed Framework: CMX}
\subsection{Framework Overview}
\label{sec:framework_overview}
The overview of CMX is shown in Fig.~\ref{fig:framework}a.
We use two parallel branches to extract features from RGB- and X-modal inputs, which can be RGB-Depth, -Thermal, -Polarization, -Event, -LiDAR data, \etc. Specifically, our proposed framework for RGB-X semantic segmentation adopts a two-branch design to effectively extract features from both RGB and X modal inputs. The two branches involve the simultaneous processing of RGB and X modal data in a parallel but interactive manner, each of which is designed to capture the unique characteristics of the respective input modality. We introduce a rectification mechanism between both branches, enabling the feature from one modality to be rectified based on the feature from another modality. Additionally, we facilitate cross-modal feature interaction by exchanging rectified features from both modalities at each stage of the two-branch architecture. Based on two-branch architecture, our framework leverages the complementary information of both modalities to enhance the performance of RGB-X semantic segmentation.

While features from different modalities have their specific noisy measurements, the feature of another modality has the potential for rectifying and calibrating the noisy information.
As shown in Fig.~\ref{fig:framework}b, we design a Cross-Modal Feature Rectification Module (\emph{CM-FRM}) to rectify one feature regarding another feature, and vice versa. In this manner, features from both modalities can be rectified. 
Besides, CM-FRMs are assembled between two adjacent stages of backbones. In this way, both rectified features are sent to the next stage to further deepen and improve the feature extraction. {Furthermore, as shown in Fig.~\ref{fig:framework}c, we design a two-stage Feature Fusion Module (\emph{FFM}) to fuse features belonging to the same level into a single feature map. Then, a decoder is used to predict the final semantic map.}
In Sec.~\ref{sec:cm_frm} and Sec.~\ref{sec:ffm}, we detail the design of CM-FRM and FFM, respectively.
In the following, we use $\mathbf{X}$ to refer to the supplementary modality, which can be Depth-, Thermal-, Polarization-, Event-, LiDAR data, \etc.

\subsection{Cross-Modal Feature Rectification}
\label{sec:cm_frm}
As analyzed above, the information originating from different sensing modalities are usually complementary~\cite{hu2019acnet,chen2020sa_gate} but contain noisy measurements.
The noisy information can be filtered and calibrated by using features coming from another modality.
{To this purpose, in Fig.~\ref{fig:framework}b, we propose a novel Cross-Modal Feature Rectification Module (\emph{CM-FRM}) to perform feature rectification between parallel streams at each stage in feature extraction.} To tackle noises and uncertainties in diverse modalities, CM-FRM processes features in two dimensions, including \emph{channel-wise} and \emph{spatial-wise} feature rectification, which together offer a holistic calibration, enabling better multi-modal feature extraction and interaction.

\noindent\textbf{Channel-wise feature rectification.}
{We embed bi-modal features $\mathbf{RGB}_{in}\in\mathbb{R}^{H \times W \times C}$ and $\mathbf{X}_{in}\in\mathbb{R}^{H \times W \times C}$ along the spatial axis into two attention vectors $\mathbf{W}_{RGB}^{C}\in\mathbb{R}^{C}$ and $\mathbf{W}_{X}^{C}\in\mathbb{R}^{C}$.
Different from previous channel-wise attention methods~\cite{chen2020sa_gate,deng2021feanet,chen2017sca}, we apply both global max pooling and global average pooling to $\mathbf{RGB}_{in}$ and $\mathbf{X}_{in}$ along the channel dimension to retain more information.
We concatenate the four resulted vectors, having $\mathbf{Y}\in\mathbb{R}^{4C}$.
Then, an MLP is applied, followed by a sigmoid function to obtain $\mathbf{W}^{C} \in \mathbb{R}^{2C}$ from ${\mathbf{Y}}$, which will be split into $\mathbf{W}_{RGB}^{C}$ and $\mathbf{W}_{X}^{C}$:
\begin{equation}
\mathbf{W}_{RGB}^{C}, \mathbf{W}_{X}^{C} = \mathcal{F}_{split}\Bigg(\sigma\bigg(\mathcal{F}_{mlp}(\rm\mathbf Y)\bigg)\Bigg),
\end{equation}
where $\sigma(\cdot)$ denotes the sigmoid function. The channel-wise rectification is then operated as:
\begin{equation}
\begin{aligned}
\mathbf{RGB}_{rec}^{C} &= \mathbf{W}_{X}^{C} \circledast \mathbf{X}_{in},\\
\mathbf{X}_{rec}^{C} &= \mathbf{W}_{RGB}^{C} \circledast \mathbf{RGB}_{in},
\end{aligned}
\end{equation}
where $\circledast$ denotes channel-wise multiplication.}

\noindent\textbf{Spatial-wise feature rectification.}
As the aforementioned channel-wise feature rectification module concentrates on learning global weights for a global calibration, we further introduce a spatial-wise feature rectification for calibrating local information.
The bi-modal inputs $\mathbf{RGB}_{in}$ and $\mathbf{X}_{in}$ will be concatenated and embedded into two spatial weight maps: $\mathbf{W}_{RGB}^{S}{\in}\mathbb{R}^{H \times W}$ and $\mathbf{W}_{X}^{S}{\in}\mathbb{R}^{H \times W}$.
The embedding operation has two $1{\times}1$ convolution layers assembled with a RELU function. Afterward, a sigmoid function is applied to obtain the embedded feature map $\mathbf{F}{\in}\mathbb{R}^{ H \times W \times 2}$, which is further split into two weight maps.
The process to obtain the spatial weight maps is formulated as:
\begin{equation}
\rm{\mathbf{F}} = {Conv}_{1\times1}\Bigg({RELU}\bigg({Conv}_{1\times1}(\mathbf{RGB}_{in} \parallel \mathbf{X}_{in})\bigg)\Bigg),
\end{equation}
\begin{equation}
\mathbf{W}_{RGB}^{S}, \mathbf{W}_{X}^{S} = \mathcal{F}_{split}\bigg(\sigma(\rm\mathbf F)\bigg).
\end{equation}
Similar to channel-wise rectification, spatial-wise rectification is formulated as:
\begin{equation}
\begin{aligned}
\mathbf{RGB}_{rec}^{S} &= \mathbf{W}_{X}^{S}  *  \mathbf{X}_{in},\\
\mathbf{X}_{rec}^{S} &= \mathbf{W}_{RGB}^{S}  * \mathbf{RGB}_{in},
\end{aligned}
\end{equation}
where $*$ denotes spatial-wise multiplication.

The whole rectified feature for both modalities $\mathbf{RGB}_{out}$ and $\mathbf{X}_{out}$ is organized as:
\begin{equation}~\label{eq:CM_FRM}
\begin{aligned}
\mathbf{RGB}_{out} &= \mathbf{RGB}_{in} + \lambda_{C}\mathbf{RGB}_{rec}^{C} + \lambda_{S}\mathbf{RGB}_{rec}^{S},\protect\\
\mathbf{X}_{out} &= \mathbf{X}_{in} + \lambda_{C}\mathbf{X}_{rec}^{C} + \lambda_{S}\mathbf{X}_{rec}^{S}.
\end{aligned}
\end{equation}
$\lambda_{C}$ and $\lambda_{S}$ are two hyperparameters. We set them both as $0.5$ as default and will ablate in Sec.~\ref{sec:ablation_study}. $\mathbf{RGB}_{out}$ and $\mathbf{X}_{out}$ are the rectified features after the comprehensive calibration, which will be sent into the next stage for feature fusion.

\subsection{Feature Fusion}
\label{sec:ffm}
After obtaining the feature maps at each layer, we build a two-stage Feature Fusion Module (\emph{FFM}) to enhance the information interaction and combination. As shown in Fig.~\ref{fig:framework}(c), in the information exchange stage (Stage $1$), the two branches are still maintained, and a cross-attention mechanism is designed to globally exchange information between the two branches.
In the fusion stage (Stage $2$), the concatenated feature is transformed into the original size via a mixed channel embedding. 

\noindent\textbf{Information exchange stage.}
At this stage, the bi-modal features will exchange their information via a symmetric dual-path structure.
For brevity, we take the X-modal path for illustration. We first flatten the input feature with size $\mathbb{R}^{H \times W \times C}$ to $\mathbb{R}^{N \times C}$, where $N{=}H{\times}W$. 
Afterward, a linear embedding is used to generate two vectors with the same size $\mathbb{R}^{N \times C_{i}}$, which we call residual vector $\mathbf{{X}^{res}}$ and interactive vector $\mathbf{{X}^{inter}}$.
We further put forward an efficient cross-attention mechanism applied to these two interactive vectors from different modal paths, which will carry out sufficient information exchange across modalities. This offers complementary interactions from the sequence-to-sequence perspective beyond the rectification-based interactions from the feature map perspective in CM-FRM.

Our cross-attention mechanism for enhancing cross-modal feature fusion is based on the traditional self-attention~\cite{vaswani2017attention}.
The original self-attention operation encodes the input vectors into Query ($\mathbf{Q}$), Key ($\mathbf{K}$), and Value ($\mathbf{V}$). The global attention map is calculated via a matrix multiplication $\mathbf{Q}\mathbf{K^{T}}$, which has a size of $\mathbb{R}^{N \times N}$ and causes a high memory occupation.
In contrast, \cite{shen2021efficient} uses a global context vector $\rm\mathbf{G} = \rm\mathbf{K^{T}V}$ with a size $\mathbb{R}^{C_{head} \times C_{head}}$ and the attention result is calculated by $\rm\mathbf{QG}$.
We flexibly {adapt} the reformulation and develop our multi-head cross-attention based on this efficient self-attention mechanism.
Specifically, the interactive vectors will be embedded into $\rm\mathbf{K}$ and $\rm\mathbf{V}$ for each head, and both sizes of them are $\mathbb{R}^{N \times C_{head}}$.
The output is obtained by multiplying the interactive vector and the context vector from the other modality path, namely a cross-attention process, and it is depicted in the following equations:
\begin{equation}
\begin{aligned}
\mathbf{G}_{RGB} &= \mathbf{K}_{RGB}^{T}\mathbf{V}_{RGB},\\ \mathbf{G}_{X} &= \mathbf{K}_{X}^{T}\mathbf{V}_{X},
\end{aligned}
\end{equation}
\begin{equation}
\begin{aligned}
\mathbf{U}_{RGB} &= \mathbf{X}_{RGB}^{inter}\ SoftMax(\mathbf{G}_{X}),\\ 
\mathbf{U}_{X} &= \mathbf{X}_{X}^{inter}\ SoftMax(\mathbf{G}_{RGB}).
\end{aligned}
\end{equation}
Note that $\rm\mathbf{G}$ denotes the global context vector, while $\rm\mathbf{U}$ indicates the attended result.
To realize the attention from different representation subspaces, we remain the multi-head mechanism, where the number of heads matches the transformer backbone.
Then, the attended result vector $\rm\mathbf{U}$ and the residual vector $\rm\mathbf{X^{res}}$ are concatenated.
Finally, we apply a second linear embedding and resize the feature to $\mathbb{R}^{H \times W \times C}$.

\noindent\textbf{Fusion stage.}
In the second stage of FFM, precisely the fusion stage, we use a simple channel embedding to merge the two paths' features, which is realized via $1{\times}1$ convolution layers.
Further, we consider that during such a channel-wise fusion, the information of surrounding areas should also be exploited for robust RGB-X segmentation. 
Thereby, inspired by Mix-FFN in~\cite{xie2021segformer} and ConvMLP~\cite{li2021convmlp}, we add one more depth-wise convolution layer ${DWConv}_{3\times3}$ to realize a skip-connected structure.
In this way, the merged features with the size $\mathbb{R}^{H \times W \times 2C}$ are fused into the final output with the size of $\mathbb{R}^{H \times W \times C}$ for feature decoding.
 
\subsection{Multi-modal Data Representations}
\label{sec:multimodal_data_representations}

\noindent\textbf{RGB-Depth.}
Depth images naturally offer range, position, and contour information. {The fusion of RGB and depth information can better separate objects with indistinguishable colors and textures at different spatial locations. We encode the depth images into HHA format~\cite{gupta2014learning}.} HHA offers geometric properties, including horizontal disparity, height above ground, and angle. 

\noindent\textbf{RGB-Thermal.}
At night or in places with insufficient light, objects, and backgrounds have similar color information and are difficult to distinguish. Thermal images provide infrared characteristics of objects, which are the potential to improve objects with thermal properties such as \emph{people}. We directly use the infrared thermal image and copy the single-channel thermal image input $3$ times to match the backbone input.

\noindent\textbf{RGB-Polarization.}
{High-reflectivity objects such as \emph{glasses} and \emph{cars} in RGB images are easily confused with surroundings. Polarization cameras record the optical polarimetric information when polarized reflection occurs, which offers complementary information in scenes with specular surfaces.} The polarization sensor is equipped with a polarization mask layer with four different directions~\cite{xiang2021polarization} and thereby each captured image set consists of four pixel-aligned images at different polarization angles $[I_{0^\circ}, I_{45^\circ}, I_{90^\circ}, I_{135^\circ}]$, where $I_{angle}$ denotes the image recorded at the corresponding angle.

We investigate two representations, \ie, the Degree of Linear Polarization ($DoLP$) and the Angle of Linear Polarization ($AoLP$), which are key polarimetric properties characterizing light polarization patterns~\cite{xiang2021polarization}.
{They are derived by Stokes vectors $S{=}
\{S_0,S_1,S_2,S_3\}$ that describe the polarization state of light.} Precisely, $S_0$ represents the total light intensity, $S_1$ and $S_2$ denote the ratio of $0^\circ$ and $45^\circ$ linear polarization over its perpendicular polarized portion, and $S_3$ stands for the circular polarization power which is not involved in our work. 
{The Stokes vectors $S_0,S_1,S_2$ can be calculated from image intensity measurements $\{I_{0^\circ}, I_{45^\circ}, I_{90^\circ}, I_{135^\circ}\}$ via:}
\begin{equation}
\begin{aligned}
S_0 &= I_{0^\circ}+I_{90^\circ}=I_{45^\circ}+I_{135^\circ},\\
S_1 &= I_{0^\circ}-I_{90^\circ},\\
S_2 &= I_{45^\circ}-I_{135^\circ}.
\end{aligned}
\end{equation}
Then, $DoLP$ and $AoLP$ are formally computed as:
\begin{equation}
DoLP = \frac{\sqrt{S_1^2+S_2^2}}{S_0},
\end{equation}
\begin{equation}
AoLP = \frac{1}{2}arctan\bigg(\frac{S_2}{S_1}\bigg).
\end{equation}
In our experiments, we further study monochromatic and trichromatic polarization cues, coupled with RGB images in multi-modal RGB-P semantic segmentation. 
For monochromatic representation used in previous works~\cite{xiang2021polarization,yan2021nlfnet}, we obtain it from monochromatic intensity measurements and convert it to $3$-channel input by copying the single-channel information.
For trichromatic polarization representation in either $DoLP$ or $AoLP$, we compute separately for their respective RGB channels.

\begin{figure}[t]
\centering
\includegraphics[width=\columnwidth]{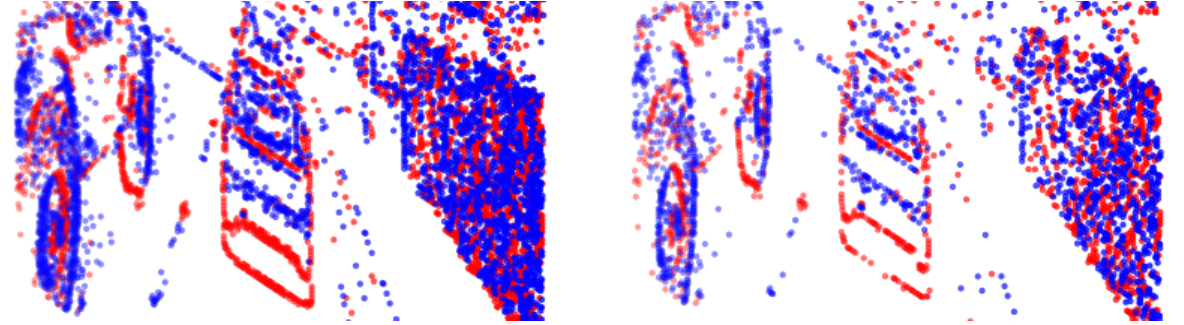}
    \begin{minipage}[t]{.5\columnwidth}\vskip -3ex
        \subcaption{Direct}\label{fig:event_pre_1}
    \end{minipage}%
    \begin{minipage}[t]{.5\columnwidth}\vskip -3ex
        \subcaption{Ours}\label{fig:event_pre_2}
    \end{minipage}%
\caption{Comparison between event representations.}
\label{fig:event_pre}
\end{figure}

\noindent\textbf{RGB-Event.}
Event data provide multiple advantages such as high dynamic range, high temporal resolution, and not being influenced by motion blur~\cite{alonso2019ev_segnet}, which are critical in dynamic scenes with motion information such as road-driving environments~\cite{zhang2021issafe,zhang2021edcnet}. 
To process event data, a set of raw events in a time window $\Delta{T}{=}t_{N}{-}t_1$ is embedded into a voxel grid with spatial dimensions $H{\times}W$ and time bins $B$, where $t_1$ and $t_N$ are the start- and the end time stamp. Unlike previous work~\cite{gehrig2021eventscape_dataset} converting event data to $B{=}3$, in this work, events are first embedded into a voxel grid with a higher time resolution, which we set the upscale size of the event bin as $6$. Then, every $6$ panels are superimposed to obtain a fine-grained event embedding.
A comparison between the \textit{direct} representation~\cite{gehrig2021eventscape_dataset} and our event representation is shown in Fig.~\ref{fig:event_pre}, in which our representation is more fine-grained in each event panel.
Apart from $B{=}3$, we further investigate different settings of event time bin $B{=}\{1,5,10,15,20,30\}$ in our method for reaching robust RGB-E semantic segmentation.

\noindent\textbf{RGB-LiDAR.}
LiDAR camera can provide reliable and accurate {spatial-depth} information on the physical world~\cite{zhuang2021pmf}. To make the representation of LiDAR data consistent with RGB images, we follow~\cite{zhuang2021pmf} to convert LiDAR data to a range-view image-like format. The Field-of-View (FoV) of the camera is $90^\circ$ and the image resolution is $H{\times}W{=}1408{\times}376$. The origin is $(u_0, v_0){=}(H/2, W/2)$. Then, the focal length $(f_x, f_y)$ can be calculated through:
\begin{equation}\label{eq:fov}
\begin{aligned}
    f_x &= H/(2{\times}tan(FoV{\times}\pi/360)), \\
    f_y &= W/(2{\times}tan(FoV{\times}\pi/360)).
\end{aligned}
\end{equation}
{Similar to~\cite{mohammadbagher2020real}, we project the LiDAR 3D points from the world coordinate to the 2D image coordinate by using:
\begin{align}
\begin{bmatrix}
u \\
v \\
1 \\
\end{bmatrix} = 
\begin{bmatrix}
f_x & 0 & u_0 & 0 \\
0 & f_y & v_0 & 0 \\
0 & 0 & 1 & 0 \\
\end{bmatrix} 
\begin{bmatrix}
\boldsymbol{R} & \boldsymbol{t} \\
\boldsymbol{0}^T_{3{\times}1} & 1 \\
\end{bmatrix}
\begin{bmatrix}
X \\
Y \\
Z \\
1 \\
\end{bmatrix},
\end{align}
where $(X,Y,Z)$ is the LiDAR point, $(u, v)$ is the 2D image pixel, and the rotation ($\boldsymbol{R}$) and the translation ($\boldsymbol{t}$) matrices are given by KITTI-360 dataset~\cite{liao2021kitti360}.}

\section{Experiment Datasets and Setups}
\subsection{Datasets}
We use five RGB-Depth semantic segmentation datasets, and datasets of RGB-Thermal, RGB-Polarization, RGB-Event, and RGB-LiDAR combinations to verify our proposed CMX.

\noindent\textbf{NYU Depth V2 dataset}~\cite{silberman2012nyu_dataset} contains $1449$ RGB-D images with the size $640{\times}480$, divided into $795$ training images and $654$ testing images with annotations on $40$ semantic categories.

\noindent\textbf{SUN-RGBD dataset}~\cite{song2015sun_rgbd} has $10335$ RGB-D images with $37$ classes, and $5285/5050$ for training/testing. Following~\cite{chen2020sa_gate,zhang2021non_aggregation}, we randomly crop and resize the input to $480{\times}480$.

\noindent\textbf{Stanford2D3D dataset}~\cite{armeni2017stanford2d3d} has $70496$ RGB-D images with $13$ object categories. Following the data splitting~\cite{cao2021shapeconv,wang2018depth_aware}, areas of $\{1, 2, 3, 4, 6\}$ are used for training and area $5$ is for testing. The input image is resized to $480{\times}480$.

\noindent\textbf{ScanNetV2 dataset}~\cite{dai2017scannet} provides $19466/5436/2135$ RGB-D samples for training/validation/testing. There are $20$ classes. %
During training, the RGB images are re-scaled to the same size of $640{\times}480$ as the depth images. During testing, the predictions are in the original size of $1296{\times}968$.

\noindent\textbf{Cityscapes dataset}~\cite{cordts2016cityscapes} is an outdoor RGB-D dataset of urban road-driving street scenes. It is divided into $2975$/$500$/$1525$ images in the training/validation/testing splits, both with finely annotated dense labels on $19$ classes. The image scenes cover $50$ different cities with a full resolution of $2048{\times}1024$.

\noindent\textbf{RGB-T MFNet dataset}~\cite{ha2017mfnet} is a multi-spectral RGB-Thermal image dataset, which has $1569$ images annotated in $8$ classes at the resolution of $640{\times}480$. $820$ images are captured during the day and the other $749$ are at night. The training set has $50\%$ of the daytime- and $50\%$ of the nighttime images, while the validation- and test set respectively have $25\%$ of the daytime- and $25\%$ of the nighttime images.

\noindent\textbf{RGB-P ZJU dataset}~\cite{xiang2021polarization} is an RGB-Polarization dataset collected by a multi-modal vision sensor designed for automated driving~\cite{sun2019multimodal} on complex campus street scenes. It is composed of $344$ images for training and $50$ images for evaluation, both labeled with $8$ semantic classes at the pixel level. The input image is resized to $612{\times}512$.

\noindent\textbf{RGB-E EventScape dataset.}
A large-scale multi-modal RGB-Event semantic segmentation benchmark is not available. To fill this gap, we create an RGB-Event multi-modal semantic segmentation benchmark\footnote[1]{\url{https://paperswithcode.com/sota/semantic-segmentation-on-eventscape}} based on the EventScape dataset~\cite{gehrig2021eventscape_dataset}, which is originally designed for depth estimation. The comparison between three event-based semantic segmentation datasets is presented in Table.~\ref{tab:event_datasets}. Unlike previous datasets using gray-scale images and pseudo labels, the RGB and the synthetic labels are available in our benchmark, which can provide more sufficient information and more precise annotations. To maintain data diversity from the original sequences generated by CARLA simulator~\cite{dosovitskiy2017carla}, we select one frame from every $30$ frames, obtaining $4077/749$ images from $122329/22493$ for training/evaluation. The images have a $512{\times}256$ resolution and are annotated with $12$ semantic classes, including \texttt{Vehicle}, \texttt{Building}, \texttt{Wall}, \texttt{Vegetation}, \texttt{Road}, \texttt{Pole}, \texttt{RoadLines}, \texttt{Fences}, \texttt{Pedestrian}, \texttt{TrafficSign}, \texttt{Sidewalk}, and \texttt{TrafficLight}.

\begin{table}[!t]
\centering
\caption{\textsc{Comparison of event-based semantic segmentation datasets.} 
}
\label{tab:event_datasets}
\resizebox{\columnwidth}{!}{   
\begin{tabular}{llllllr}
\toprule
Dataset       & Image      & Event & Train/Val  & Label     & Resolution & \multicolumn{1}{l}{Class} \\
\midrule\midrule
DDD17~\cite{alonso2019ev_segnet}         & Gray-scale & 50Hz  & 15950/3890 & pseudo    & 346 $\times$ 260  & 6 \\
DSEC-Semantic~\cite{sun2022ess} & Gray-scale        & 20Hz & 8082/2809  & pseudo    & 640 $\times$ 440  & 11 \\
EventScape~\cite{gehrig2021eventscape_dataset}    & RGB        & 500Hz & 122329/22493   & synthetic & 512 $\times$ 256  & 12 \\
\bottomrule
\end{tabular}
}
\end{table}

\noindent\textbf{RGB-L KITTI-360 dataset.}
KITTI-360~\cite{liao2021kitti360} is a suburban driving dataset, which has $49004/12276$ images at the size of $1408{\times}376$ for training/validation. There are $19$ semantic classes following the Cityscapes dataset~\cite{cordts2016cityscapes}.

\subsection{Implementation Details}
During training on all datasets, data augmentation is performed by random flipping and scaling with random scales $[0.5,1.75]$.
We take Mix Transformer encoder (MiT) pre-trained on ImageNet~\cite{russakovsky2015imagenet} as the backbone and MLP-decoder with an embedding dimension of $512$ unless specified, both introduced in SegFormer~\cite{xie2021segformer}. We select AdamW optimizer~\cite{kingma2014adam} with weight decay $0.01$.
The original learning rate is set as $6e^{-5}$ and we employ a poly learning rate schedule.
We use cross-entropy as the loss function.
When reporting multi-scale testing results on NYU Depth V2 and SUN RGB-D, we use multiple scales $\{0.75,1,1.25\}$ with horizontal flipping.
We use mean Intersection over Union (mIoU) averaged across semantic classes as the primary evaluation metric to measure the segmentation performance.
More specific settings for different datasets are described in detail in the appendix.

\section{Experimental Results and Analyses}
In this section, we present experimental results to verify the effectiveness of our proposed CMX for RGB-X semantic segmentation.
In Sec.~\ref{sec:results_rgbd}, we show the results of CMX on multiple indoor and outdoor RGB-Depth benchmarks, compared with state-of-the-art methods.
In Sec.~\ref{sec:results_rgbt}, we analyze the RGB-Thermal segmentation performance for robust daytime- and nighttime semantic perception.
In Sec.~\ref{sec:results_rgbp} and Sec.~\ref{sec:results_rgbe}, we study the generalization of CMX to RGB-Polarization and RGB-Event modality combinations and representations of these multi-modal data.
In Sec.~\ref{sec:results_rgbl}, we present the results of CMX on the RGB-LiDAR dataset.
In Sec.~\ref{sec:ablation_study}, we conduct a comprehensive variety of ablation studies to confirm the effects of different components in our solution.
Finally, we perform efficiency- and qualitative analysis in Sec.~\ref{sec:efficiency_analysis} and Sec.~\ref{sec:qualitative_analysis}.
\begin{table*}[!t]
\centering
\caption{\textsc{Results on five RGB-Depth datasets. \textit{Acc} and $^{*}$ denote pixel accuracy and multi-scale test.}}
\vskip -1ex
\label{tab:rbgd_sota}
    \begin{subtable}[t]{0.6\columnwidth}
    \begin{subtable}[t]{\columnwidth}
    \caption{Results on NYU Depth V2~\cite{silberman2012nyu_dataset}.}
    \vskip -1ex
    \label{tab:NYU}
    \setlength{\tabcolsep}{2pt}
    \resizebox{\columnwidth}{!}{    
    \renewcommand{\arraystretch}{0.93}
    	\begin{tabular}{l|cc}
        \toprule
        \textbf{Method} & \textbf{mIoU (\%) }  & \textbf{Acc (\%)} \\
        \midrule\midrule
        3DGNN~\cite{qi20173d} & 43.1 & - \\
        Kong $et\ al.$~\cite{kong2018recurrent} & 44.5 & 72.1 \\
        LS-DeconvNet~\cite{cheng2017locality} & 45.9 & 71.9\\
        CFN~\cite{CaRF17} & 47.7 & - \\
        ACNet~\cite{hu2019acnet} & 48.3 & - \\
        RDF-101~\cite{park2017rdfnet} & 49.1 & 75.6\\
        SGNet~\cite{chen2021spatial_guided} & 51.1 & 76.8\\
        ShapeConv~\cite{cao2021shapeconv} & 51.3 & 76.4 \\
        NANet~\cite{zhang2021non_aggregation} & 52.3 & 77.9 \\
        SA-Gate~\cite{chen2020sa_gate} & 52.4 & 77.9 \\
        \midrule
        CMX (MiT-B2) & 54.1 & 78.7 \\
        CMX (MiT-B2)$^{*}$ & \textbf{54.4} & \textbf{79.9} \\
        CMX (MiT-B4) & 56.0 & 79.6 \\
        CMX (MiT-B4)$^{*}$ & \textbf{56.3} & \textbf{79.9} \\
        CMX (MiT-B5) & 56.8 & 79.9 \\
        \rowcolor{gray!15} CMX (MiT-B5)$^{*}$ & \textbf{56.9} & \textbf{80.1} \\
        \bottomrule
        \end{tabular}
    }
    \end{subtable}%
    \hspace{\fill}
    \begin{subtable}[t]{\columnwidth}
    \centering
    \vskip 2ex
    \caption{Results on Stanford2D3D~\cite{armeni2017stanford2d3d}.}
    \vskip -1ex
    \label{tab:table_stanford2d3d}
    \resizebox{\columnwidth}{!}{
    \renewcommand{\arraystretch}{1.}
    \setlength{\tabcolsep}{1pt}
        \begin{tabular}{l|cc}
        \toprule
        \textbf{Method} & \textbf{mIoU (\%) }  & \textbf{Acc (\%)}\\
        \midrule\midrule
        Depth-aware CNN~\cite{wang2018depth_aware} & 39.5 & 65.4 \\
        MMAF-Net-152~\cite{fooladgar2019mmafnet} & 52.9 & 76.5 \\
        ShapeConv-101~\cite{cao2021shapeconv} & 60.6 & \textbf{82.7} \\
        \midrule
        CMX (MiT-B2) & 61.2 & 82.3 \\ 
        \rowcolor{gray!15} CMX (MiT-B4) & \textbf{62.1} & 82.6 \\
        \bottomrule
        \end{tabular}
    }
    \end{subtable}
    \end{subtable}
    \hspace{\fill}
    \begin{subtable}[t]{0.6\columnwidth}
    \begin{subtable}[t]{\columnwidth}
    \caption{Results on SUN-RGBD~\cite{song2015sun_rgbd}.}
    \vskip -1ex
    \label{tab:table_SUN}
    \setlength{\tabcolsep}{4pt}
    \resizebox{\columnwidth}{!}{    
    \renewcommand{\arraystretch}{0.9}
    	\begin{tabular}{l|cc}
        \toprule
        \textbf{Method} & \textbf{mIoU (\%) }  & \textbf{Acc (\%)}\\
        \midrule\midrule
        3DGNN~\cite{qi20173d} & 45.9  & - \\
        RDF-152~\cite{park2017rdfnet} & 47.7 & 81.5 \\
        CFN~\cite{CaRF17}  & 48.1 & - \\
        D-CNN~\cite{wang2018depth_aware} & 42.0 & - \\
        ACNet~\cite{hu2019acnet} & 48.1  & - \\
        TCD~\cite{yue2021two_stage} & 49.5 & 83.1\\
        SGNet~\cite{chen2021spatial_guided}  & 48.6  & 82.0 \\
        SA-Gate~\cite{chen2020sa_gate} & 49.4  & 82.5 \\
        NANet~\cite{zhang2021non_aggregation} & 48.8 & 82.3\\
        ShapeConv~\cite{cao2021shapeconv} & 48.6  & 82.2\\
        \midrule
        CMX (MiT-B2)$^{*}$ & 49.7 & 82.8 \\ 
        CMX (MiT-B4)$^{*}$ & 52.1 & 83.5\\
        \rowcolor{gray!15} CMX (MiT-B5)$^{*}$ & \textbf{52.4} & \textbf{83.8}\\
        \bottomrule
        \end{tabular}
    }
    \end{subtable}%
    \hspace{\fill}
    \begin{subtable}[t]{\columnwidth}
    \centering
    \vskip 2ex
    \caption{Results on ScanNetV2 \textit{test} set~\cite{dai2017scannet}.
    }
    \vskip -1ex
    \label{tab:table_scannet}
    \resizebox{\columnwidth}{!}{
    \renewcommand{\arraystretch}{0.98}
    \setlength{\tabcolsep}{4pt}{
        \begin{tabular}{l|cc}
        \toprule
        \textbf{Method} & \textbf{Modal} & \textbf{mIoU (\%) } \\
        \midrule\midrule
        PSPNet~\cite{zhao2017pspnet} & RGB & 47.5  \\
        AdapNet++~\cite{valada2019ssma} & RGB & 50.3 \\
        \midrule
        3DMV (2d-proj)~\cite{dai20183dmv} & RGB-D & 49.8 \\
        FuseNet~\cite{hazirbas2016fusenet} & RGB-D & 53.5 \\
        SSMA~\cite{valada2019ssma} & RGB-D & 57.7 \\
        GRBNet~\cite{qian2021rfbnet_gated} & RGB-D & 59.2 \\
        MCA-Net~\cite{shi2020mcanet} & RGB-D & 59.5 \\
        DMMF~\cite{shi2022dmmf_label_oriented} & RGB-D & 59.7\\
        \midrule
        \rowcolor{gray!15} CMX (MiT-B2) & RGB-D & \textbf{61.3}\\ 
        \bottomrule
        \end{tabular}
    }
    }
    \end{subtable}
    \end{subtable}
    \hspace{\fill}
    \begin{subtable}[t]{0.7\columnwidth}
    \caption{Results on Cityscapes \emph{val} set~\cite{cordts2016cityscapes}.}
    \vskip -1ex
    \label{tab:table_CS}
    \setlength{\tabcolsep}{1pt}
    \resizebox{\columnwidth}{!}{    
    \renewcommand{\arraystretch}{1.40}
        \begin{tabular}{l|cc|c}
        \toprule
        \textbf{Method}& \textbf{Modal} & \textbf{Backbone} & \textbf{mIoU (\%) } \\
        \midrule\midrule
        SwiftNet~\cite{orsic2019swiftnet} & RGB & ResNet-18 & 70.4 \\
        ESANet~\cite{seichter2020efficient} & RGB & ResNet-50 & 79.2 \\
        GSCNN~\cite{takikawa2019gated_scnn} &RGB &WideResNet-38& 80.8 \\
        CCNet~\cite{huang2019ccnet} &RGB &ResNet-101& 81.3 \\
        DANet~\cite{fu2019danet} &RGB&ResNet-101& 81.5 \\
        ACFNet~\cite{zhang2019acfnet}&RGB&ResNet-101 & 81.5 \\
        SegFormer~\cite{xie2021segformer}& RGB & MiT-B2 & 81.0 \\
        SegFormer~\cite{xie2021segformer}& RGB & MiT-B4 & 82.3 \\
        \midrule
        RFNet~\cite{sun2020rfnet} & RGB-D & ResNet-18 & 72.5 \\
        PADNet~\cite{xu2018padnet} &RGB-D & ResNet-50& 76.1 \\
        Kong $et\ al.$~\cite{kong2018recurrent}&RGB-D&ResNet-101 & 79.1 \\
        ESANet~\cite{seichter2020efficient} & RGB-D & ResNet-50 & 80.0 \\
        SA-Gate~\cite{chen2020sa_gate}  &RGB-D& ResNet-50 & 80.7 \\
        SA-Gate~\cite{chen2020sa_gate}  &RGB-D& ResNet-101& 81.7 \\
        AsymFusion~\cite{wang2020channel_shuffle} &RGB-D& Xception65 & 82.1 \\
        SSMA~\cite{valada2019ssma} &RGB-D& ResNet-50 & 82.2 \\
        \midrule
        CMX & RGB-D & MiT-B2 & \textbf{81.6} \\
        \rowcolor{gray!15} CMX & RGB-D & MiT-B4 & \textbf{82.6} \\
        \bottomrule
        \end{tabular}
    }
    \end{subtable}%
\end{table*}

\subsection{Results on RGB-Depth Datasets}
\label{sec:results_rgbd}
We first conduct experiments on RGB-D semantic segmentation datasets. The results are grouped in Table~\ref{tab:rbgd_sota}.

\noindent\textbf{NYU Depth V2.}
The results on the NYU Depth V2 dataset are shown in Table~\ref{tab:NYU}. It can be easily seen that our approach achieves leading scores.
The proposed method with MiT-B2 already exceeds previous methods, attaining $54.4\%$ in mIoU.
{Our CMX models based on MiT-B4 and -B5 further dramatically improve the mIoU to $56.3\%$ and $56.9\%$, clearly standing out in front of all state-of-the-art approaches.}
The best CMX model even reaches superior results than recent strong pretraining-based methods~\cite{girdhar2022omnivore,bachmann2022multimae} like Omnivore~\cite{girdhar2022omnivore} that uses images, videos, and single-view 3D data for supervision.

\noindent\textbf{Stanford2D3D.} {In Table~\ref{tab:table_stanford2d3d}, our CMX achieves state-of-the-art mIoU scores. Our B2-based CMX surpasses the previous best ShapeConv~\cite{cao2021shapeconv} based on ResNet-101~\cite{he2016resnet} in mIoU} and our model based on MiT-B4 further reaches mIoU of $62.1\%$. The results demonstrate the effectiveness and learning capacity of our approach on such a large RGB-D dataset.

\noindent\textbf{SUN-RGBD.} 
As presented in Table~\ref{tab:table_SUN}, our method achieves leading performances on the SUN-RGBD dataset. Our interactive cross-modal fusion approach (Fig.~\ref{fig2_3:interact_fusion}) exceeds previous input fusion methods (Fig.~\ref{fig2_1:input_fusion}), \eg, SGNet~\cite{chen2021spatial_guided} and ShapeConv~\cite{cao2021shapeconv}, as well as feature fusion methods (Fig.~\ref{fig2_2:feat_fusion}), \eg, ACNet~\cite{hu2019acnet} and SA-Gate~\cite{chen2020sa_gate}.
In particular, with MiT-B4 and -B5, CMX elevates the mIoU to ${>}52.0\%$.
CMX is also better than multi-task methods like PAP~\cite{zhang2019pattern} and TET~\cite{zhang2022tube_embedded}.

\noindent\textbf{ScanNetV2.}
We test our CMX model with MiT-B2 on the ScanNetV2 benchmark.
As shown in Table~\ref{tab:table_scannet}, it can be clearly seen that CMX outperforms RGB-only methods and achieves the top mIoU of $61.3\%$ among the RGB-D methods.
On the ScanNetV2 leaderboard, methods like BPNet~\cite{hu2021bidirectional_projection} reach higher scores by using 3D supervision from point clouds to perform joint 2D- and 3D reasoning. In contrast, our method attains a competitively accurate performance by using purely 2D data and effectively leveraging the complementary information inside RGB-D modalities.

\noindent\textbf{Cityscapes.}
Besides indoor RGB-D datasets, to study the generalizability to outdoor scenes, we assess the effectiveness of CMX on Cityscapes. 
As shown in Table~\ref{tab:table_CS}, we note that the improvement on the Cityscapes dataset is not as obvious as other datasets, because the performance of RGB-only models on this dataset shows a saturation trend.
Compared with MiT-B2 (RGB), our RGB-D approach elevates the mIoU by $0.6\%$.
Our approach based on MiT-B4 achieves a state-of-the-art score of $82.6\%$, outstripping all existing RGB-D methods by more than $0.4\%$ in absolute mIoU values, verifying that CMX generalizes well to street scene understanding.

\subsection{Results on RGB-Thermal Dataset}
\label{sec:results_rgbt}

\begin{table*}[!t]
    \begin{center}
        \caption{\textsc{Per-class results on MFNet dataset~\cite{ha2017mfnet} for RGB-Thermal segmentation.}}
        \label{tab:rgbt_sota_result}
        \begin{tabular}{ l | c | c c c c c c c c c | c }
    \toprule[1pt]
    \textbf{Method} & \textbf{Modal} & \textbf{Unlabeled} & \textbf{Car} & \textbf{Person} & \textbf{Bike} & \textbf{Curve} & \textbf{Car Stop} & \textbf{Guardrail} & \textbf{Color Cone} & \textbf{Bump} & \textbf{mIoU} \\
    \midrule\midrule
    ERFNet~\cite{romera2018erfnet} & RGB & 96.7 & 67.1 & 56.2 & 34.3 & 30.6 & 9.4 & 0.0 & 0.1 & 30.5 & 36.1 \\
    DANet~\cite{fu2019danet} & RGB & 96.3 & 71.3 & 48.1 & 51.8 & 30.2 & 18.2 & 0.7 & 30.3 & 18.8 & 41.3 \\
    PSPNet~\cite{zhao2017pspnet} & RGB & 96.8 & 74.8 & 61.3 & 50.2 & 38.4 & 15.8 & 0.0 & 33.2 & 44.4 & 46.1\\ 
    HRNet~\cite{wang2020hrnet} & RGB & 98.0 & 86.9 & 67.3 & 59.2 & 35.3 & 23.1 & 1.7 & 46.6 & 47.3 & 51.7 \\
    SegFormer-B2~\cite{xie2021segformer} & RGB & 97.9 & 87.4 & 62.8 & 63.2 & 31.7 & 25.6 & 9.8 & 50.9 & 49.6 & 53.2 \\
    SegFormer-B4~\cite{xie2021segformer} & RGB & 98.0 & 88.9 & 64.0 & 62.8 & 38.1 & 25.9 & 6.9 & 50.8 & 57.7 & 54.8 \\
    \midrule
    MFNet~\cite{ha2017mfnet} & RGB-T & 96.9 & 65.9 & 58.9 & 42.9 & 29.9 & 9.9 & 0.0 & 25.2 & 27.7 & 39.7 \\
    SA-Gate~\cite{chen2020sa_gate}& RGB-T & 96.8 & 73.8 & 59.2 & 51.3 & 38.4 & 19.3 & 0.0 & 24.5 & 48.8 & 45.8\\
    Depth-aware CNN~\cite{wang2018depth_aware} & RGB-T & 96.9 & 77.0 & 53.4 & 56.5 & 30.9 & 29.3 & 8.5 & 30.1 & 32.3 & 46.1 \\
    ACNet~\cite{hu2019acnet} & RGB-T & 96.7 & 79.4 & 64.7 & 52.7 & 32.9 & 28.4 & 0.8 & 16.9 & 44.4 & 46.3\\
    PSTNet~\cite{shivakumar2020pst900} & RGB-T & 97.0 & 76.8 & 52.6 & 55.3 & 29.6 & 25.1 & \textbf{15.1} & 39.4 & 45.0 & 48.4 \\
    RTFNet~\cite{sun2019rtfnet} & RGB-T & 98.5 & 87.4 & 70.3 & 62.7 & 45.3 & 29.8 & 0.0 & 29.1 & 55.7 & 53.2 \\
    FuseSeg~\cite{sun2020fuseseg} & RGB-T & 97.6 & 87.9 & 71.7 & 64.6 & 44.8 & 22.7 & 6.4 & 46.9 & 47.9 & 54.5 \\
    AFNet~\cite{xu2021attention_fusion} & RGB-T & 98.0 & 86.0 & 67.4 & 62.0 & 43.0 & 28.9 & 4.6 & 44.9 & 56.6 & 54.6\\
    ABMDRNet~\cite{zhang2021abmdrnet} & RGB-T & \textbf{98.6} & 84.8 & 69.6 & 60.3 & 45.1 & 33.1 & 5.1 & 47.4 & 50.0 & 54.8\\
    FEANet~\cite{deng2021feanet} & RGB-T & 98.3 & 87.8 & 71.1 & 61.1 & 46.5 & 22.1 & 6.6 & \textbf{55.3} & 48.9 & 55.3\\
    DHFNet~\cite{cai2023dhfnet} & RGB-T & 97.7 & 87.6 & 71.7 & 61.1 & 39.5 & \textbf{42.4} &  9.5 & 49.3 & 56.0 & 57.2\\
    GMNet~\cite{zhou2021gmnet} & RGB-T & 97.5 & 86.5 & 73.1 & 61.7 & 44.0 & 42.3 & 14.5 & 48.7 & 47.4 & 57.3\\
    \midrule
    \rowcolor{gray!15} CMX (MiT-B2) & RGB-T & 98.3 & 89.4 & 74.8 & \textbf{64.7} & 47.3 & 30.1 & 8.1 & 52.4 & 59.4 & 58.2\\
   \rowcolor{gray!15} CMX (MiT-B4) & RGB-T & 98.3 & \textbf{90.1} & \textbf{75.2} & 64.5 & \textbf{50.2} & {35.3} & 8.5 & 54.2 & \textbf{60.6} & \textbf{59.7} \\
    \bottomrule[1pt]
\end{tabular}
    \end{center}
\end{table*}
\noindent\textbf{Comparison with the state-of-the-art.}
In Table~\ref{tab:rgbt_sota_result}, we compare our method against RGB-only models and multi-modal methods using RGB-T inputs of MFNet dataset~\cite{ha2017mfnet}.
As unfolded, ACNet~\cite{hu2019acnet} and SA-Gate~\cite{chen2020sa_gate}, carefully designed for RGB-Depth segmentation, perform less satisfactorily on RGB-T data, as they focus on feature extraction without sufficient feature interaction before fusion and thereby fail to generalize to other modality. Depth-aware CNN~\cite{wang2018depth_aware}, an input fusion method with modality-specific operator design, also does not yield high performance.
In contrast, the proposed CMX strategy, enabling comprehensive interactions from various perspectives, generalizes smoothly in RGB-T semantic segmentation.
It can be seen that our method based on MiT-B2 achieves mIoU of $58.2\%$, clearly outperforming the previous best RGB-T methods ABMDRNet~\cite{zhang2021abmdrnet}, FEANet~\cite{deng2021feanet}, and GMNet~\cite{zhou2021gmnet}.
Our CMX with MiT-B4 further elevates state-of-the-art mIoU to $59.7\%$, widening the accuracy gap in contrast to existing methods.
Moreover, it is worth pointing out that the improvements brought by our RGB-X approach compared with the RGB-only baselines are compelling, \ie, ${+}5.0\%$ and ${+}4.9\%$ in mIoU for MiT-B2 and -B4 backbones, respectively.
Our approach overall achieves top scores on \emph{car}, \emph{person}, \emph{bike}, \emph{curve}, \emph{car stop}, and \emph{bump}. 
For \emph{person} with infrared properties, our approach enjoys more than ${+}11.0\%$ gain in IoU, confirming the effectiveness of CMX in harvesting complementary cross-modal information.

\begin{table}[t]
    \begin{center}
        \caption{\textsc{Segmentation results on daytime- and nighttime images on MFNet dataset~\cite{ha2017mfnet}.}}
        \label{tab:rgbt_daynight_result}
        
\resizebox{\columnwidth}{!}{    
\begin{tabular}{l | c | c | c}
    \toprule[1pt]
    \textbf{Method} & \textbf{Modal} & \textbf{Daytime mIoU (\%)} & \textbf{Nighttime mIoU (\%)}\\
    \midrule\midrule
    FRRN~\cite{pohlen2017full} & RGB & 40.0 & 37.3 \\
    DFN~\cite{yu2018dfn} & RGB & 38.0 & 42.3\\
    BiSeNet~\cite{yu2018bisenet} & RGB & 44.8 & 47.7 \\
    SegFormer-B2~\cite{xie2021segformer} & RGB & 48.6 & 49.2\\
    SegFormer-B4~\cite{xie2021segformer} & RGB & 49.4 & 52.4 \\
    \midrule
    MFNet~\cite{ha2017mfnet} & RGB-T & 36.1 & 36.8\\
    FuseNet~\cite{hazirbas2016fusenet} & RGB-T & 41.0 & 43.9\\
    RTFNet~\cite{sun2019rtfnet} & RGB-T & 45.8 & 54.8\\
    FuseSeg~\cite{sun2020fuseseg} & RGB-T & 47.8 & 54.6\\
    GMNet~\cite{zhou2021gmnet} & RGB-T & 49.0 & 57.7\\
    \midrule
    \rowcolor{gray!15} CMX (MiT-B2) & RGB-T & 51.3 & 57.8\\
    \rowcolor{gray!15} CMX (MiT-B4) & RGB-T & \textbf{52.5} & \textbf{59.4}\\
    \bottomrule[1pt]
\end{tabular}
}
    \end{center}
\end{table}
\noindent\textbf{Day and night performances.}
Following \cite{sun2020fuseseg,zhou2021gmnet}, we assess day- and night segmentation results on the RGB-T benchmark (see Table~\ref{tab:rgbt_daynight_result}).
For daytime scenes, our approach increases mIoU by $2.7\%{\sim}3.1\%$ compared with RGB-only baselines. At nighttime, RGB segmentation often suffers from poor lighting conditions, and it even carries much noisy information in the RGB data. Yet, our CMX rectifies the noisy images and exploits supplementary features from thermal data, dramatically improving the mIoU by ${>}7.0\%$ and enhancing the robustness of semantic scene understanding in unfavorable environments with adverse illuminations.

\subsection{Results on RGB-Polarization Dataset}
\label{sec:results_rgbp}

\begin{table*}[!t]
    \begin{center}
        \caption{\textsc{Per-class results on ZJU-RGB-P~\cite{xiang2021polarization} dataset for RGB-Polarization segmentation.}}
        \label{tab:per_class_rgbp}
        \begin{tabular}{ l | c |c c c c c c c c | c}
    \toprule[1pt]
    \textbf{Method} & \textbf{Modal} & {\textbf{Building}} & {\textbf{Glass}} & {\textbf{Car}} & {\textbf{Road}} & {\textbf{Vegetation}} & {\textbf{Sky}} & {\textbf{Pedestrian}} & {\textbf{Bicycle}} & \textbf{mIoU}\\
    \midrule\midrule
    SwiftNet~\cite{orsic2019swiftnet} & RGB & 83.0 & 73.4 & 91.6 & 96.7 & 94.5 & 84.7 & 36.1 & 82.5 & 80.3\\
    SegFormer-B2~\cite{xie2021segformer} & RGB & 90.6 & 79.0 & 92.8 & 96.6 & 96.2 & 89.6 & 82.9 & 89.3 & 89.6\\\midrule
    NLFNet~\cite{yan2021nlfnet} & RGB-P & 85.4 & 77.1 & 93.5 & 97.7 & 93.2 & 85.9 & 56.9 & 85.5 & 84.4\\
    EAFNet~\cite{xiang2021polarization} & RGB-P & 87.0 & 79.3 & 93.6 & 97.4 & 95.3 & 87.1 & 60.4 & 85.6 & 85.7\\ 
    \midrule
    CMX (SegFormer-B2) & RGB-AoLP (Monochromatic) & \textbf{91.9} & 87.0 & 95.6 & 98.2 & 96.7 & 89.0 & 84.9 & 92.0 & 91.8\\
    CMX (SegFormer-B2) & RGB-AoLP (Trichromatic) & 91.5 & 87.3 & 95.8 & 98.2 & 96.6 & 89.3 & 85.6 & 91.9 & 92.0\\
    CMX (SegFormer-B4) & RGB-AoLP (Monochromatic)& 91.8 & \textbf{88.8} & \textbf{96.3} & \textbf{98.3} & 96.7 & 89.1 & 86.3 & 92.3 & 92.4\\
    \rowcolor{gray!15} CMX (SegFormer-B4) & RGB-AoLP (Trichromatic) & 91.6 & \textbf{88.8} & \textbf{96.3} & \textbf{98.3} & \textbf{96.8} & 89.7 & 86.2 & \textbf{92.8} & \textbf{92.6}\\
    \midrule
    CMX (SegFormer-B2) & RGB-DoLP (Monochromatic) & 91.4 & 87.6 & 96.0 & 98.2 & 96.6 & 89.1 & 87.1 & 92.3 & 92.1\\
    CMX (SegFormer-B2) & RGB-DoLP (Trichromatic) & 91.8 & 87.8 & 96.1 & 98.2 & 96.7 & \textbf{89.4} & 86.1 & 91.8 & 92.2\\
    CMX (SegFormer-B4) & RGB-DoLP (Monochromatic) & 91.8 & 88.6 & \textbf{96.3} & \textbf{98.3} & 96.7 & 89.4 & 86.0 & 92.1 & 92.4\\
    \rowcolor{gray!15} CMX (SegFormer-B4) & RGB-DoLP (Trichromatic) & 91.6 & 88.6 & \textbf{96.3} & \textbf{98.3} & 96.7 & 89.5 & \textbf{86.4} & 92.2 & 92.5\\
    \bottomrule[1pt]
\end{tabular}
    \end{center}
\end{table*}
\noindent\textbf{Comparison with the state-of-the-art.}
Table~\ref{tab:per_class_rgbp} shows per-class accuracy of our approach compared to RGB-only~\cite{xie2021segformer,orsic2019swiftnet} and RGB-Polarization fusion methods~\cite{xiang2021polarization,yan2021nlfnet} on ZJU-RGB-P dataset~\cite{xiang2021polarization}.
Our unified CMX outperforms the previous best RGB-P method~\cite{xiang2021polarization} by ${>}6.0\%$ in mIoU.
We observe that the improvement on \emph{pedestrian} is significant thanks to the capacity of the transformer backbone and our cross-modal fusion mechanisms.
Compared to the RGB-only baseline with MiT-B2~\cite{xie2021segformer}), the IoU improvements on classes with polarimetric characteristics are clear, such as \emph{glass} (${>}8.0\%$) and \emph{car} (${>}2.5\%$), further evidencing the generalizability of our cross-modal fusion solution in bridging RGB-P streams.

\noindent\textbf{Analysis of polarization data representations.}
We study polarimetric data representations and the results displayed in Table~\ref{tab:per_class_rgbp} indicate that the Angle of Linear Polarization ($AoLP$) and the Degree of Linear Polarization ($DoLP$) representations both carry effective polarization information beneficial for semantic scene understanding, which is consistent with the finding in~\cite{xiang2021polarization}.
Besides, trichromatic representations are consistently better than monochromatic representations used in previous RGB-P segmentation works~\cite{xiang2021polarization,yan2021nlfnet}.
This is expected as the trichromatic representation provides more detailed information, which should be leveraged to fully unlock the potential of trichromatic polarization cameras.

\begin{table}[!t]
    \begin{center}
        \caption{\textsc{Results for RGB-Event segmentation.}}
        \label{tab:rgb_event_result}
        
\setlength{\tabcolsep}{2pt}
\begin{tabular}{l|cc|cc}
    \toprule[1pt]
    \textbf{Method} & \textbf{Modal} & \textbf{Backbone} & \textbf{mIoU (\%)}  & \textbf{Pixel  Acc. (\%)}\\
    \midrule\midrule
    SwiftNet~\cite{orsic2019swiftnet} & RGB & ResNet-18 & 36.67 & 83.46 \\
    Fast-SCNN~\cite{Poudel2019FastSCNNFS} & RGB & Fast-SCNN & 44.27 & 87.10\\
    CGNet~\cite{Wu2021CGNetAL}  & RGB & M3N21 & 44.75 & 87.13\\
    Trans4Trans~\cite{zhang2021trans4trans_iccvw} & RGB & PVT-B2 & 51.86 & 89.03 \\ 
    Swin-s~\cite{liu2021swin} & RGB & Swin-s & 52.49 & 88.78 \\
    Swin-b~\cite{liu2021swin} & RGB & Swin-b & 53.31 & 89.21 \\
    DeepLabV3+~\cite{chen2018deeplabv3+} & RGB & ResNet-101 & 53.65 & 89.92 \\
    SegFormer-B2~\cite{xie2021segformer} & RGB & MiT-B2 & 58.69 & 91.21 \\
    SegFormer-B4~\cite{xie2021segformer} & RGB & MiT-B4 & 59.86 & 91.61 \\

    \midrule
    RFNet~\cite{sun2020rfnet} & RGB-E & ResNet-18 & 41.34 & 86.25 \\
    ISSAFE~\cite{zhang2021issafe} & RGB-E & ResNet-18 & 43.61 & 86.83 \\
    SA-Gate~\cite{chen2020sa_gate} & RGB-E & ResNet-101 & 53.94 & 90.03 \\ 

    \midrule
    CMX (DeepLabV3+) & RGB-E & ResNet-101 & 54.91 & 89.67\\
    CMX (Swin-s) & RGB-E & Swin-s & 60.86 & 91.25 \\
    CMX (Swin-b) & RGB-E & Swin-b & 61.21 & 91.61 \\
    CMX (SegFormer-B2) & RGB-E & MiT-B2 & 61.90 & 91.88 \\
    \rowcolor{gray!15} CMX (SegFormer-B4) & RGB-E & MiT-B4 & \textbf{64.28} & \textbf{92.60} \\
    \bottomrule
\end{tabular}
    \end{center}
\end{table}
\subsection{Results on RGB-Event Dataset}
\label{sec:results_rgbe}

\begin{figure}[t]
   \centering
   \includegraphics[width=1.0\linewidth, keepaspectratio]{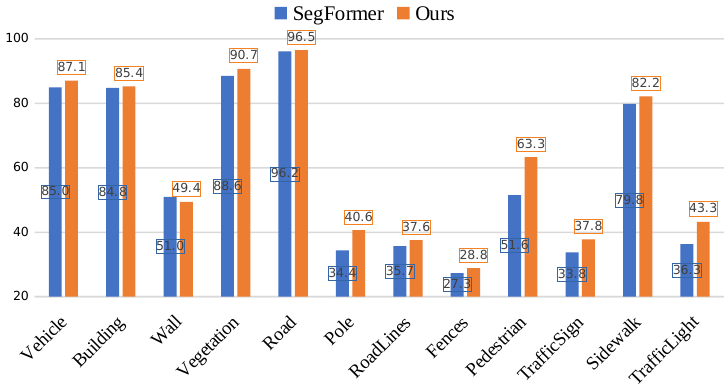}
   \caption{Per-class IoU results of the RGB-only baseline and our RGB-Event model on our RGB-Event benchmark.} 
   \label{fig:RGBE_perclass}
\end{figure}

\noindent\textbf{Comparison with the state-of-the-art.}
In Table~\ref{tab:rgb_event_result}, we benchmark more than $10$ semantic segmentation methods, including RGB-only methods,  CNN-based~\cite{orsic2019swiftnet,Poudel2019FastSCNNFS,Wu2021CGNetAL,chen2018deeplabv3+} and transformer-based~\cite{liu2021swin,xie2021segformer,zhang2021trans4trans_iccvw} methods, as well as multi-modal methods~\cite{sun2020rfnet,chen2020sa_gate,zhang2021issafe}. In contrast, our models improve performance by mixing RGB-Event features, as seen in Table~\ref{tab:rgb_event_result} and Fig.~\ref{fig:RGBE_perclass}.
Our model using MiT-B4 reaches $64.28\%$ in mIoU, towering over all other methods and setting the state-of-the-art on the RGB-E benchmark.
This further verifies the versatility of our solution for different multi-modal combinations.
Fig.~\ref{fig:RGBE_perclass} depicts a per-class accuracy comparison between the RGB baseline and our RGB-Event model with MiT-B2.
With event data, the foreground objects are more accurately parsed by our RGB-E model, \eg, \emph{vehicle} (${+}2.1\%$), \emph{pedestrian} (${+}11.7\%$), and traffic light (${+}7.0\%$).

\noindent\textbf{Analysis of using different backbones.}
To verify that our unified method is effective with using different backbones, we compare CNN- and transformer-based backbones in the CMX framework. Specifically, in addition to MiT backbones, we experiment with DeepLabV3+~\cite{chen2018deeplabv3+} and Swin transformer~\cite{liu2021swin} backbones with UperNet~\cite{xiao2018upernet} to construct CMX.
Compared to the RGB-only DeepLabV3+, Swin-s, and Swin-b methods, CMX models achieve respective ${+}1.26\%$, ${+}8.37\%$, ${+}7.90\%$ gains in mIoU. The results show that our RGB-X solution consistently improves the segmentation performance, confirming that our unified framework is not strictly tied to a concrete backbone type, but can be flexibly deployed with CNN- or transformer models, which helps to yield effective unified architecture for RGB-X semantic segmentation.

\begin{figure}[!t]
    \begin{center}
       \includegraphics[width=0.9\linewidth, keepaspectratio]{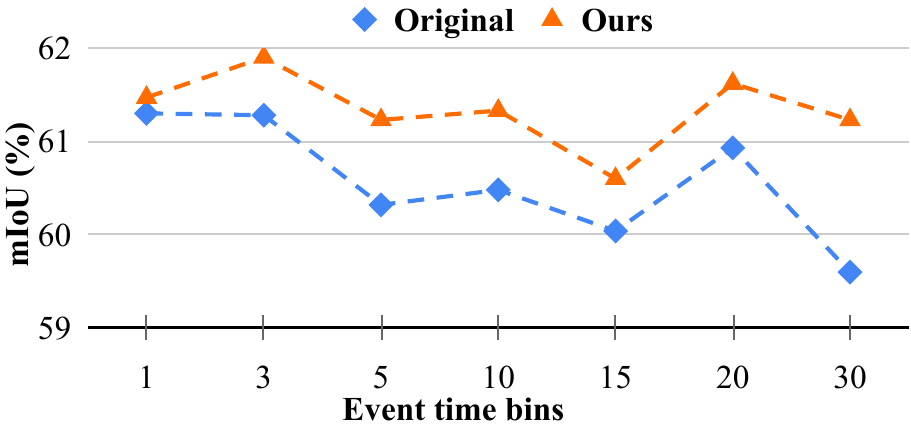}
    \end{center}
       \caption{
       Analysis of event representations and time bins.
       } 
    \label{fig:rgbe_timebins_chart}
\end{figure}

\noindent\textbf{Analysis of event data representations.}
We study with different settings of event time bin $B{=}\{1,3,5,10,15,20,30\}$ 
based on our CMX fusion model with MiT-B2. Compared with the original event representation~\cite{gehrig2021eventscape_dataset}, our representation achieves consistent improvements (in Fig.~\ref{fig:rgbe_timebins_chart}) on different settings of event time bins, such as ${+}1.63\%$ of mIoU when $B{=}30$.
In particular, it helps our CMX to obtain the highest mIoU of $61.90\%$ in the setting of $B{=}3$.
In $B{=}1$, embedding all events in a single time bin leads to dragging behind images of moving objects and being sub-optimal for feature fusion.
In higher time bins, events produced in a short interval are dispersed to more bins, resulting in insufficient events in a single bin.
These corroborate observations in~\cite{zhang2021issafe,zhang2021edcnet} and that the event representation $B{=}3$ is an effective time bin setting for RGB-E semantic segmentation with CMX.

\subsection{Results on RGB-LiDAR Dataset}\label{sec:results_rgbl}
\begin{table}[t]
\centering
\caption{\textsc{Results for RGB-LiDAR segmentation.}}    
\label{tab:rgb_lidar_result}
\resizebox{\columnwidth}{!}{
\setlength{\tabcolsep}{10pt}{
	\begin{tabular}{l|c|c}
    \toprule
    \textbf{Method} & \textbf{Backbone} & \textbf{mIoU (\%)} \\
    \midrule\midrule
    HRFuser~\cite{broedermann2022hrfuser}& HRFormer-T &48.74\\
    PMF~\cite{zhuang2021pmf} & SalsaNext & 54.48\\
    TokenFusion~\cite{wang2022tokenfusion} & MiT-B2 & 54.55\\
    TransFuser~\cite{prakash2021transfuser} & RegNetY-3.2GF & 56.57\\
    \rowcolor{gray!15} CMX & MiT-B2 & \textbf{64.31}\\
    \bottomrule
    \end{tabular}
}
}
\end{table}
In Table~\ref{tab:rgb_lidar_result}, we compare CMX with other models dedicated to RGB-LiDAR data fusion, including PMF~\cite{zhuang2021pmf} and TransFuser~\cite{prakash2021transfuser}. These two methods achieve respective $54.48\%$ and $56.57\%$ in mIoU. Besides, other general multimodal fusion methods, \eg, HRFuser~\cite{broedermann2022hrfuser} and TokenFusion~\cite{wang2022tokenfusion}, are included for comparison. In contrast, our CMX obtains the state-of-the-art performance with $64.31\%$ in mIoU, having a ${+}9.76\%$ gain compared with TokenFusion which is also based on MiT-B2. The sufficient improvement proves the advantage of using a symmetric dual-stream architecture in modal fusion and the effectiveness of our proposed cross-modal rectification and fusion methods.

\subsection{Ablation Study}
\label{sec:ablation_study}
We perform a series of ablation studies to explore how different parts of our architecture affect the segmentation.
We use depth information encoded into HHA as the complementary modality here.
We take MiT-B2 as the backbone with the MLP decoder in our ablation studies unless specified.
The semantic segmentation performance is evaluated on NYU Depth V2.

\noindent\textbf{RGB-only Baseline and CMX.}
{In order to comprehensively compare the RGB-only baseline~\cite{xie2021segformer} and our RGB-X-based model, we conduct experiments on five different types of modality fusion, including RGB-Depth, -Thermal, -Polarization, -Event, and -LiDAR. Both methods are based on the same backbone with MiT-B2~\cite{xie2021segformer}  As presented in Table~\ref{tab:ablation_baseline}, on six different datasets, \ie, NYU Depth V2, Cityscapes, MFNet, ZJU-RGB-P, EventScape, and KITTI-360, our CMX model obtains improvements of ${+}6.1\%$, ${+}0.6\%$, ${+}5.0\%$, ${+}2.6\%$, ${+}3.2\%$, and ${+}3.0\%$, respectively. We note that the improvement on the Cityscapes dataset is not as obvious as other datasets, because the performance of RGB-only models on this dataset shows a saturation trend. Nonetheless, the consistent improvements achieved across five different multi-modal fusion tasks are a strong testament to the effectiveness of our proposed unified CMX framework for RGB-X semantic segmentation.}

\begin{table*}[!t]
    \begin{center}
        \caption{\textsc{Comparison between RGB-only baseline and our CMX model for RGB-X semantic segmentation, where all results (mIoU) are based on the same backbone with MiT-B2.}} 
        \label{tab:ablation_baseline}
        \resizebox{\linewidth}{!}{
        \begin{tabular}{l|l|l|l|l|l|l|l} 
\midrule
\textbf{Method} & \textbf{Modal} & \textbf{NYU Depth V2}  & \textbf{Cityscapes} & \textbf{MFNet} & \textbf{ZJU-RGB-P} & \textbf{EventScape} & \textbf{KITTI-360}\\ \midrule
SegFormer-B2~\cite{xie2021segformer} & RGB-only & 48.0  & 81.0  & 53.2 & 89.6 & 58.7 & 61.3 \\
CMX-B2 & Multimodal & 54.1 (RGB-D)& 81.6 (RGB-D) & 58.2 (RGB-T)  & 92.2 (RGB-P) & 61.9 (RGB-E) & 64.3 (RGB-L)\\
\midrule
\end{tabular}

        }
    \end{center}
\end{table*}

\noindent\textbf{Effectiveness of CM-FRM and FFM.}
We design CM-FRM and FFM to rectify and merge features coming from the RGB- and X-modality branches. We take out these two modules from the architecture respectively, where the results are shown in Table~\ref{tab:ab_study_whole}.
If CM-FRM is ablated, the features will be extracted independently in their own branches, and for FFM we simply average the two features for semantic prediction.
Compared with the baseline, using only CM-FRM improves mIoU by $2.5\%$, using only FFM improves mIoU by $1.2\%$, and together CM-FRM and FFM improve the semantic segmentation performance by $3.8\%$. The improvements show that our CM-FRM and FFM modules are both crucial for the success of the unified CMX framework.
\begin{table}[!t]
    \begin{center}
    \caption{\textsc{Ablation study of CM-FRM and FFM on NYU Depth V2 \textit{test} set. \textit{Avg.} is the average fusion.}}
    \label{tab:ab_study_whole}
    \resizebox{\columnwidth}{!}{
\setlength{\tabcolsep}{12pt}
\begin{tabular}{ll|cc}
    \toprule
    \textbf{CM-FRM} & \textbf{FFM} & \textbf{mIoU (\%)} & \textbf{Pixel Acc. (\%)} \\
    \midrule\midrule
    $\times$ & \textit{Avg.} & 50.3 & 76.8\\
    \checkmark & \textit{Avg.} & 52.8 & 78.0\\
    $\times$ & \checkmark & 51.5 & 77.1\\
    \rowcolor{gray!20}\checkmark & \checkmark & \textbf{54.1} & \textbf{78.7}\\
    \bottomrule
\end{tabular}
}
    \end{center}
\end{table}

\noindent\textbf{Ablation with CM-FRM and FFM variants.}
We further experiment with variants of CM-FRM and FFM modules.
As shown in Table~\ref{tab:ab_study_variants}, \emph{channel only} denotes using channel-wise rectification only ($\lambda_{C}{=}1$ and $\lambda_{S}{=}0$ in Eq.~\ref{eq:CM_FRM}), and \emph{spatial only} means using spatial-wise rectification only ($\lambda_{C}{=}0$ and $\lambda_{S}{=}1$ in Eq.~\ref{eq:CM_FRM}).
It can be seen that substituting the proposed CM-FRM by either \emph{channel-only} or \emph{spatial-only} variant causes a sub-optimal accuracy, further confirming the efficacy of combining the bi-modal rectification for holistic feature calibration, which is crucial for robust multi-modal segmentation.
In our channel-wise calibration, we use both global average pooling and global max pooling to retain more information. Table~\ref{tab:ab_study_variants} shows that using only global average pooling (\emph{avg. p.}) and using only global max pooling (\emph{max. p.}) are less effective than our complete CM-FRM, which offers a more comprehensive rectification. 
\begin{table}[h]
    \setlength{\tabcolsep}{2.0pt}
    \begin{center}
    \caption{\textsc{Ablation with CM-FRM/FFM variants on NYU Depth V2 \textit{test} set.}}
    \label{tab:ab_study_variants}
    \begin{tabular}{ll|cc}
    \toprule
    \textbf{Feature Rectify} & \textbf{Feature Fusion} & \textbf{mIoU (\%)} & \textbf{Pixel Acc. (\%)} \\
    \midrule\midrule
    \cellcolor{gray!10}CM-FRM \emph{channel only} & FFM & 53.6 & 78.5\\
    \cellcolor{gray!10}CM-FRM \emph{spatial only} & FFM & 53.3 & 78.3\\
    \midrule
    \cellcolor{gray!10}CM-FRM \emph{avg. p. only} & FFM & 53.0 & 78.1\\
    \cellcolor{gray!10}CM-FRM \emph{max. p. only} & FFM & 53.5 & 78.5\\
    \midrule
    CM-FRM & \cellcolor{gray!10}FFM \emph{stage 2 only} & 53.8 & 78.5\\
    CM-FRM & \cellcolor{gray!10}FFM \emph{self attn} & 53.8 & 78.6\\
    \midrule
    CM-FRM & FFM & \textbf{54.1} & \textbf{78.7}\\
    \bottomrule
\end{tabular}
    \end{center}
\end{table}

Previous ablation studies support the design of CM-FRM.
To understand the capability of FFM, we here test with two variants.
As shown in Table~\ref{tab:ab_study_variants}, \emph{stage 2 only} means there is no information exchange before the mixed channel embedding, whereas \emph{self attn} denotes that context vectors will not be exchanged in stage 1 of FFM. 
The two variants are less constructive as compared to our complete FFM. Thanks to the crucial cross-attention design for information exchange, our complete FFM productively rectifies and fuses the features at different levels. These indicate the importance of fusion from the sequence-to-sequence perspective, which is not considered in previous works.
Overall, the ablation shows that our interactive strategy, providing comprehensive interactions, is effective for cross-modal fusion.

\noindent\textbf{Ablation of the supplementary modality.}
Previous works have shown that multi-modal segmentation has a better performance than single-modal RGB segmentation~\cite{hu2019acnet}. We carry out experiments to certify that and the results are shown in Table~\ref{tab:ab_study_modalities}.
Note that here, the MLP decoder is not used, in order to focus on studying the influence of feature extraction from different supplementary modalities.
As compared to the RGB-only method, we conduct experiments with modalities of RGB-RGB, RGB-Noise, RGB-Depth, and RGB-HHA. We found that replacing the supplementary modality with random noise can obtain even better results than two RGB inputs. 
This means that even pure noise information may help the model identify noisy information in the RGB branch. The model learns to focus on relevant features and thus gains robustness. It may also help prevent over-fitting during the learning process.
However, when using depth information, we have observed obvious improvements, which further proves that the fusion of RGB and depth information brings clearly better predictions.
Encoding depth images using the HHA representation further increases the scores.
The overall gain of $5.3\%$ in mIoU, compared with the RGB-only baseline, is also compelling, which is similar to that in RGB-T semantic segmentation, demonstrating the effectiveness of our proposed method for rectifying and fusing cross-modal information.
\begin{table}[h]
\begin{center}
    \caption{\textsc{Ablation of the supplementary modality on NYU Depth V2 \textit{test} set.}}
    \label{tab:ab_study_modalities}
    \resizebox{\columnwidth}{!}{
\setlength{\tabcolsep}{18pt}
\begin{tabular}{@{\hskip 1mm}l|cc}
    \toprule
    \textbf{Modalities} & \textbf{mIoU (\%)} & \textbf{Pixel Acc. (\%)} \\
    \midrule\midrule
    RGB & 46.7 & 73.8\\\midrule
    RGB + RGB & 47.2 & 74.1\\
    RGB + Noise & 47.7 & 74.5\\
    RGB + Raw depth & \textbf{51.1} & \textbf{75.7}\\
    RGB + HHA & \textbf{52.0} & \textbf{77.0}\\
    \bottomrule
\end{tabular}
}
\end{center}
\end{table}

\subsection{Efficiency Analysis}
\label{sec:efficiency_analysis}
In Table~\ref{tab:efficiency}, we present the computational complexity results. Compared with the previous best method SA-Gate~\cite{chen2020sa_gate} on the NYU Depth V2 dataset, our model with MiT-B2 has similar \#Params and lower FLOPs but significantly higher mIoU. Our CMX model with MiT-B4 greatly elevates the mIoU score to $56.0\%$, further widening the accuracy gap with moderate model complexity.
With MiT-B5, mIoU further increases to $56.8\%$, but it also comes with larger complexity.
For efficiency-critical applications, the CMX solution with MiT-B2 or -B4 would be preferred to enable both accurate and efficient multi-modal semantic scene perception.
\begin{table}[!t]
    \begin{center}
    \caption{\textsc{Efficiency results. FLOPs are estimated for inputs of RGB and HHA, with a size of $480{\times}640{\times}3$.}}
    \label{tab:efficiency}
    \resizebox{\columnwidth}{!}{
        \begin{tabular}{l|ccc}
    \toprule
    \textbf{Method} & \textbf{\#Params (M)}  & \textbf{FLOPs (G)} & \textbf{mIoU (\%)} \\
    \midrule\midrule
    SA-Gate~\cite{chen2020sa_gate} (ResNet50) & \textbf{63.4} & 204.9 & 50.4 \\
    CMX (SegFormer-B2) & 66.6 & \textbf{67.6} & 54.1 \\
    CMX (SegFormer-B4) & 139.9 & 134.3 & 56.0 \\
    CMX (SegFormer-B5) & 181.1 & 167.8 & \textbf{56.8} \\
    \bottomrule
\end{tabular}
    }
    \end{center}
\end{table}
\subsection{Qualitative Analysis}
\label{sec:qualitative_analysis}

\begin{figure*}[!t]
    \centering
    \includegraphics[width=\linewidth]{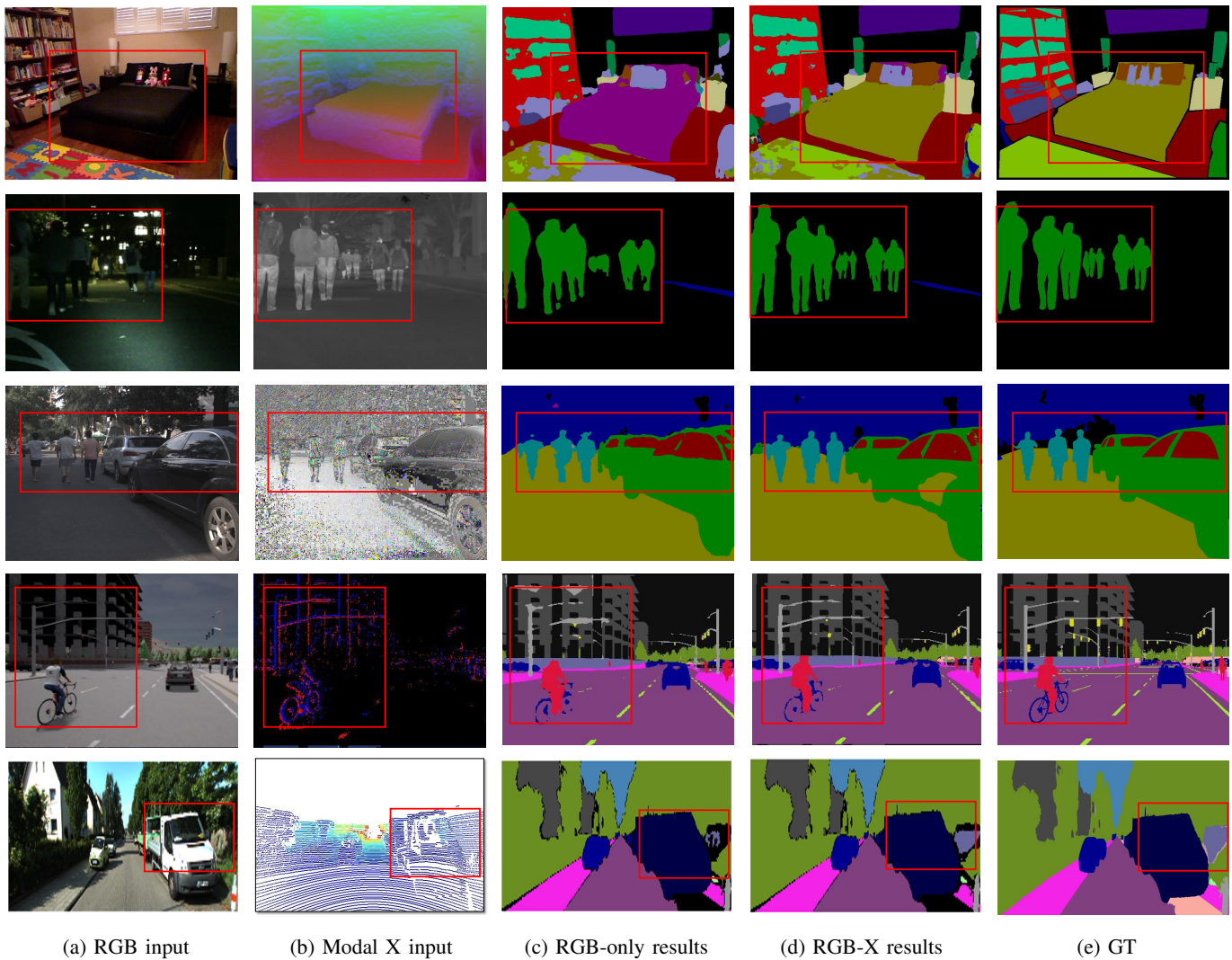}
    \begin{minipage}[t]{.2\linewidth}
        \vskip -2ex
        \subcaption{RGB input}
    \end{minipage}%
    \begin{minipage}[t]{.2\linewidth}
        \vskip -2ex
        \subcaption{Modal X input}
    \end{minipage}%
    \begin{minipage}[t]{.2\linewidth}
        \vskip -2ex
        \subcaption{RGB-only results}
    \end{minipage}%
    \begin{minipage}[t]{.2\linewidth}
        \vskip -2ex
        \subcaption{RGB-X results}
    \end{minipage}%
    \begin{minipage}[t]{.2\linewidth}
        \vskip -2ex
        \subcaption{GT}
    \end{minipage}%
    \caption{{Visualization results of RGB-only and RGB-X methods, where both are based on the same backbone. 
    From top to bottom: RGB-Depth, RGB-Thermal, RGB-Polarization (AoLP), RGB-Event, and RGB-LiDAR semantic segmentation.}}
    \label{fig:qualitative}
\end{figure*}
\noindent\textbf{Visualization of segmentation results.}
We compare the results of the RGB-only baseline and our CMX, where both are based on SegFormer-B2. We analyze each row from top to bottom in Fig.~\ref{fig:qualitative}. 
\begin{compactitem}
    \item[(1)] For RGB-Depth, we present results from the NYU Depth V2 dataset~\cite{silberman2012nyu_dataset}. CMX leverages geometric information and correctly identifies the \emph{bed} while the model wrongly classifies it as a \emph{sofa}. It proves that the CMX model can obtain discriminative features from depth information in the low-texture scenario.
    \item[(2)] For RGB-Thermal, our CMX demonstrates improvement over the baseline under low illumination conditions, \eg, the night scene. The use of Thermal in addition to RGB enables the model to make much clearer boundaries, such as between \emph{persons} and \emph{unlabeled} background. Besides, by combining features from both modalities, our CMX can more effectively filter out the noise and other unwanted artifacts that can negatively impact segmentation accuracy. For example, the segmentation of \emph{persons} in the distance is easily disturbed by overexposed lights in RGB, which can be rectified by Thermal modality. 
    \item[(3)] For RGB-Polarization, the specular \emph{glass} areas are more precisely parsed by our CMX model, as compared to the baseline. Besides, the \emph{cars} which also contain polarization cues are completely and smoothly segmented with delineated borders, and the boundaries of\emph{pedestrians} also show beneficial effects.
    \item[(4)] For RGB-Event, our CMX generalizes well and enhances the segmentation of moving objects, such as the segmentation results of \emph{cyclists} and \emph{poles}. It indicates that incorporating features extracted from Event data can enhance the modeling of dynamics that are not captured by RGB images alone.
    \item[(5)] For RGB-LiDAR, thanks to the spatial information from the LiDAR modality, our CMX model can correctly recognize the \emph{wall}, while the RGB-only method misidentifies it as part of a \emph{truck}. Furthermore, our CM-FRM module makes CMX robust against the noise of LiDAR modality, such as the \emph{truck} glass area, yielding a complete segmentation mask of the \emph{truck}.
\end{compactitem}

Overall, the qualitative examination backs up that our general approach is suitable for a diverse mix of multi-modal sensing combinations for robust semantic scene understanding.

\section{Conclusion}
To revitalize multi-modal pixel-wise semantic scene understanding for autonomous vehicles, we investigate RGB-X semantic segmentation and propose CMX, a universal transformer-based cross-modal fusion architecture, which is generalizable to a diverse mix of sensing data combinations.
We put forward a Cross-Modal Feature Rectification Module (CM-FRM) and a Feature Fusion Module (FFM) for facilitating interactions toward accurate RGB-X semantic segmentation.
CM-FRM conducts channel- and spatial-wise rectification, rendering comprehensive feature calibration.
FFM intertwines cross-attention and mixed channel embedding for enhanced global information exchange.
To further assess the generalizability of CMX to dense-sparse data fusion, we establish an RGB-Event semantic segmentation benchmark.
We study effective representations of polarimetric- and event data, indicating the optimal path to follow for reaching robust multi-modal semantic segmentation.
The proposed model sets the new state-of-the-art on nine benchmarks, spanning five RGB-D datasets, as well as RGB-Thermal, RGB-Polarization, RGB-Event, and RGB-LiDAR combinations.

\bibliographystyle{IEEEtran}
\bibliography{bib}

\clearpage

\appendices
\counterwithin{figure}{section}
\section{More Implementation Details}
\label{sec:more_implementation_details}
We implement our experiments with PyTorch. We employ a poly learning rate schedule with a factor of $0.9$ and an initial learning rate of $6e^{-5}$. The number of warm-up epochs is $10$.
We now describe implementation details for different datasets.

\noindent\textbf{NYU Depth V2 dataset.} We train our model with the MiT-B2 backbone on four 2080Ti GPUs, models with MiT-B4 and MiT-B5 backbones on three 3090 GPUs. The number of training epochs is set as $500$. We take the whole image with the size $640{\times}480$ for training and inference. We use a batch size of $8$ for the MiT-B2 backbone and $6$ for MiT-B4 and -B5.

\noindent\textbf{SUN-RGBD dataset.}
The models are trained with a batch size of $4$ per GPU.
During training, the images are randomly cropped to $480{\times}480$.
The model based on MiT-B2 is trained on two V100 GPUs for $200$ epochs. The models based on MiT-B4 and MiT-B5 are trained on eight V100 GPUs, $250$ epochs for MiT-B4 and $300$ epochs for MiT-B5.

\noindent\textbf{Stanford2D3D dataset.} The model is trained on four 2080Ti GPUs. The number of training epochs here is set as $32$. We resize the input images to $480{\times}480$. We use a batch size of $12$ for the MiT-B2 backbone and $8$ for MiT-B4.

\noindent\textbf{ScanNetV2 dataset.} The model is trained on four 2080Ti GPUs. The number of training epochs here is set as $100$. We resize the input RGB images to $640{\times}480$. We use a batch size of $12$ for the MiT-B2 backbone.

\noindent\textbf{Cityscapes dataset.} The model is trained on eight A100 GPUs for $500$ epochs. The batch size is set as $8$. The images are randomly cropped into $1024{\times}1024$ for training and inference is performed on the full resolution with a sliding window of $512{\times}512$. The embedding dimension of the MiT-B4 backbone and MLP-decoder is set as $768$.

\noindent\textbf{RGB-T MFNet dataset.} The model is trained on four 2080Ti GPUs. We use the original image size of $640\times480$ for training and inference. The batch size is set to $8$ for the MiT-B2 backbone and we train for $500$ epochs.
Consistent with the batch size of $8$, the model based on MiT-B4 is trained on four A100 GPUs, which requires a larger memory.

\noindent\textbf{RGB-P ZJU dataset.} The model is trained on four 2080Ti GPUs. We resize the image from $1224{\times}1024$ to $612{\times}512$. The number of training epochs is set as $400$. We use a batch size of $8$ for the MiT-B2 backbone and $4$ for MiT-B4. In practice, we calculate the image encoding pixel-wise $AoLP$ information by mapping the values of $arctan(S_1/S_2)$ to the range of $[0,255]$.

\noindent\textbf{RGB-E EventScape dataset.} The proposed model is trained with a batch size of $4$ and the original resolution of $512{\times}256$ on a single 1080Ti GPU. The number of training epochs is set as $100$. The embedding dimension of the MiT-B4 backbone and MLP-decoder is set as $768$.

\noindent\textbf{RGB-L KITTI-360 dataset.} The model is trained with a batch size of $2$ and the original resolution of $1408{\times}376$. The number of training epochs is set as $40$. 

\begin{figure*}[!t]
    \centering
    \includegraphics[width=\linewidth]{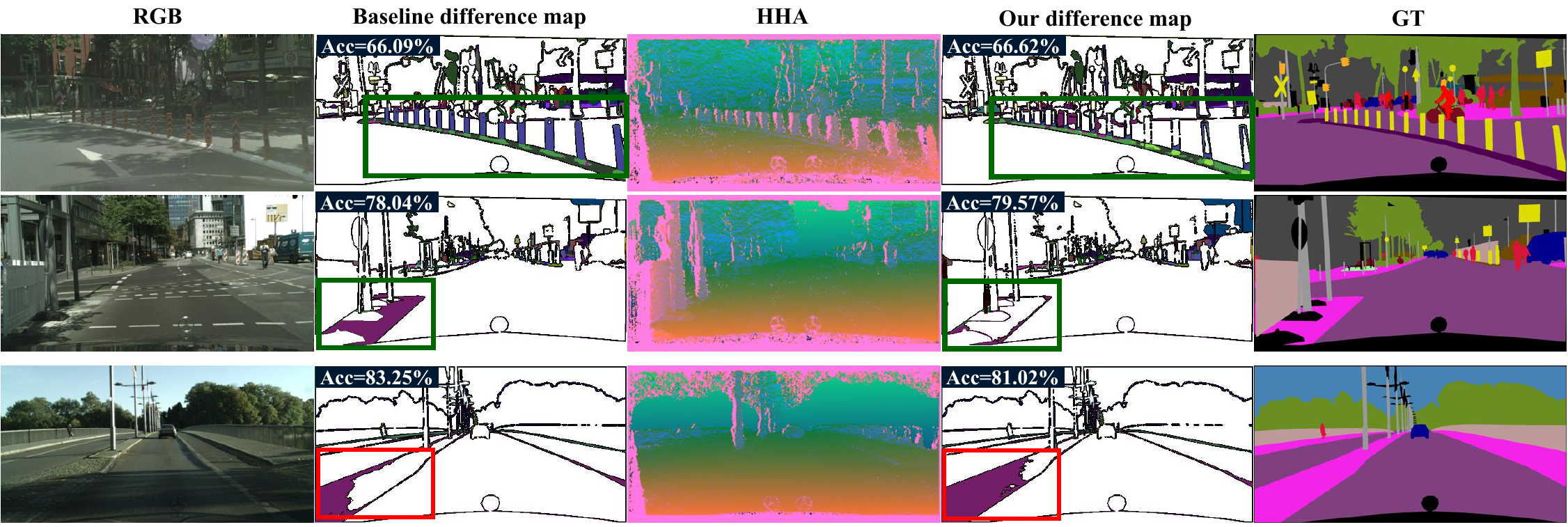}
    \caption{Visualization of semantic segmentation results for the RGB-only baseline and our RGB-X approach, both of which are based on SegFormer-B4. ``Acc'' is short for pixel accuracy of the segmentation result. From left to right: RGB image, baseline difference map \textit{w.r.t.} the ground truth, HHA image encoding depth information, our difference map, and ground truth.}
    \label{fig:cs_vis}
\end{figure*}
\begin{figure*}[!t]
    \centering
        \includegraphics[width=\linewidth]{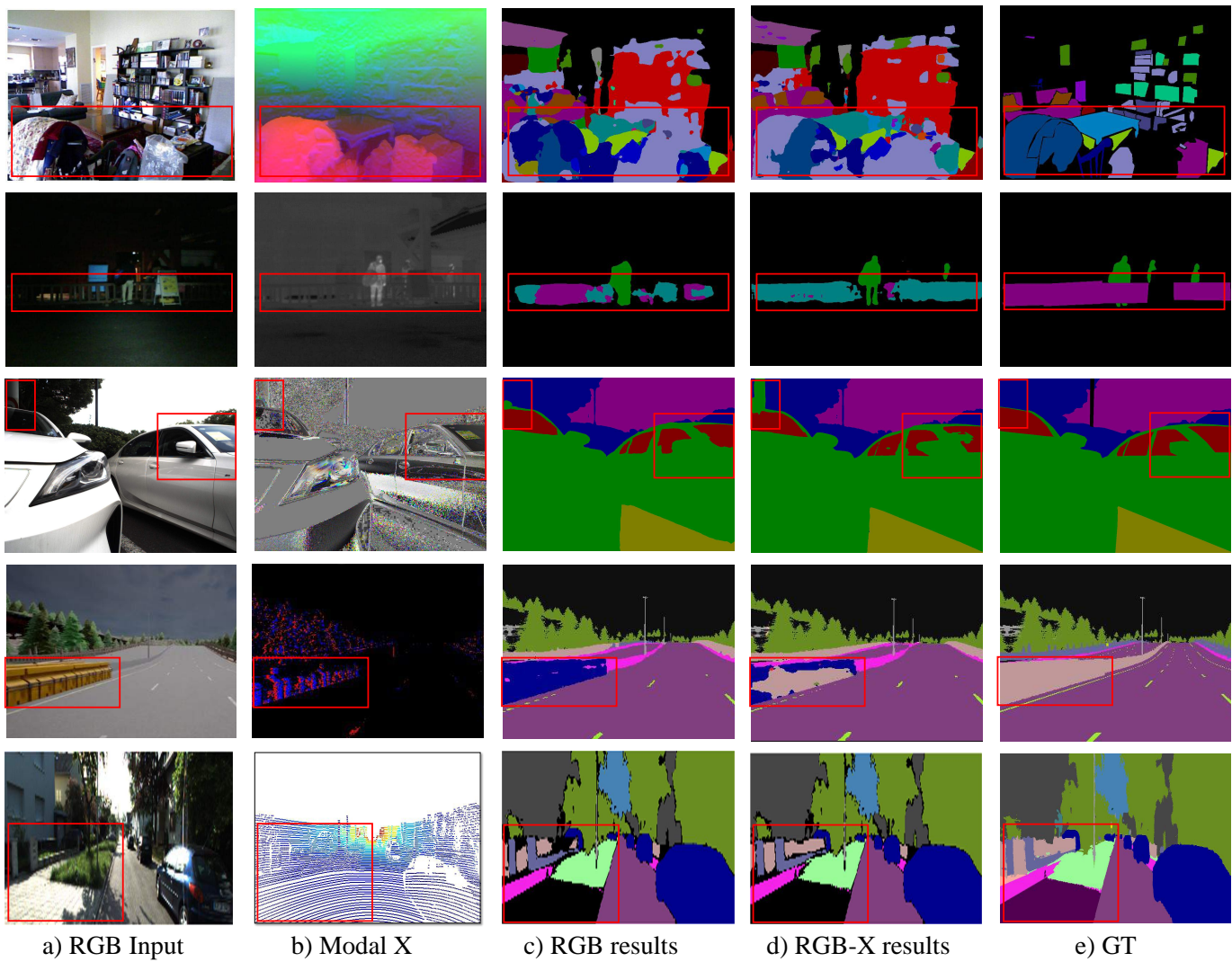}
    \caption{Visualization of failure cases. We use SegFormer-B2 for RGB segmentation and the proposed approach with the same backbone MiT-B2 and MLP-Decoder for RGB-X segmentation. From top to bottom: RGB-Depth, RGB-Thermal, RGB-Polarization (AoLP), and RGB-Event semantic segmentation.}
    \label{fig:failure_cases}
\end{figure*}

\section{More Qualitative Analysis}
\label{sec:failure_case}

\noindent\textbf{Segmentation results on the Cityscapes dataset.}
We further view the outdoor RGB-D semantic segmentation results on the Cityscapes dataset based on the backbone of SegFormer-B4.
We show the results of the RGB-only baseline and our RGB-X approach, in particular, the difference maps \textit{w.r.t.} the segmentation ground truth.
As displayed in Fig.~\ref{fig:cs_vis}, in spite of the noisy depth measurements, our CMX still benefits from the HHA-encoded image, thanks to the ability to rectify and fuse cross-modal complementary features.
Our approach has higher pixel accuracy scores on a wide variety of driving scene elements such as \emph{fence}, and \emph{sidewalk} in the positive group (in green boxes).
However, the shadows and weak illumination conditions are still challenging for both models and make the depth cues less effective. For example, depth information in the regions of \emph{sidewalk} in the negative group (in red boxes), may be less informative for fusion.

\noindent\textbf{Failure case analysis.}
In Fig.~\ref{fig:failure_cases}, we show a set of failure cases in different sensing modality combination scenarios.
The first row shows that for the RGB-D semantic segmentation in a highly composite indoor scene with extremely densely arranged objects, the parsing results are still less visually satisfactory.
In the second row of a nighttime scene, the \emph{guardrails} are misclassified by the RGB-X method as \emph{color cone}, despite our model delivering more complete and consistent segmentation than the RGB-only model and having better segmentation of \emph{person} with thermal properties.
This illustrates that at night, the perception of some remote objects is still challenging in RGB-T semantic segmentation and it should be noted for safety-critical applications like automated driving.
In the third row, the RGB-P model might be misguided by the polarized background area in an occluded situation and yields less accurate parsing results, indicating that polarization, as a strong prior for segmentation of specular surfaces like \emph{glass} and \emph{car} regions, should be carefully leveraged in unconstrained scenes with a lot of occlusions.
In the fourth row, the \emph{fences} are partially detected as \emph{vehicles} in the RGB-E segmentation result, but our model still yields more correctly identified pixels than the RGB-only model by harvesting complementary cues from event data. In the last row, the over-exposed \emph{sidewalk} region is still a challenge for segmentation. Nonetheless, our RGB-LiDAR CMX predicts a much better mask on the \emph{fence} region, where the spatial information given by LiDAR data is more accurate.

\begin{figure*}[!t]
    \centering
    \includegraphics[width=\linewidth]{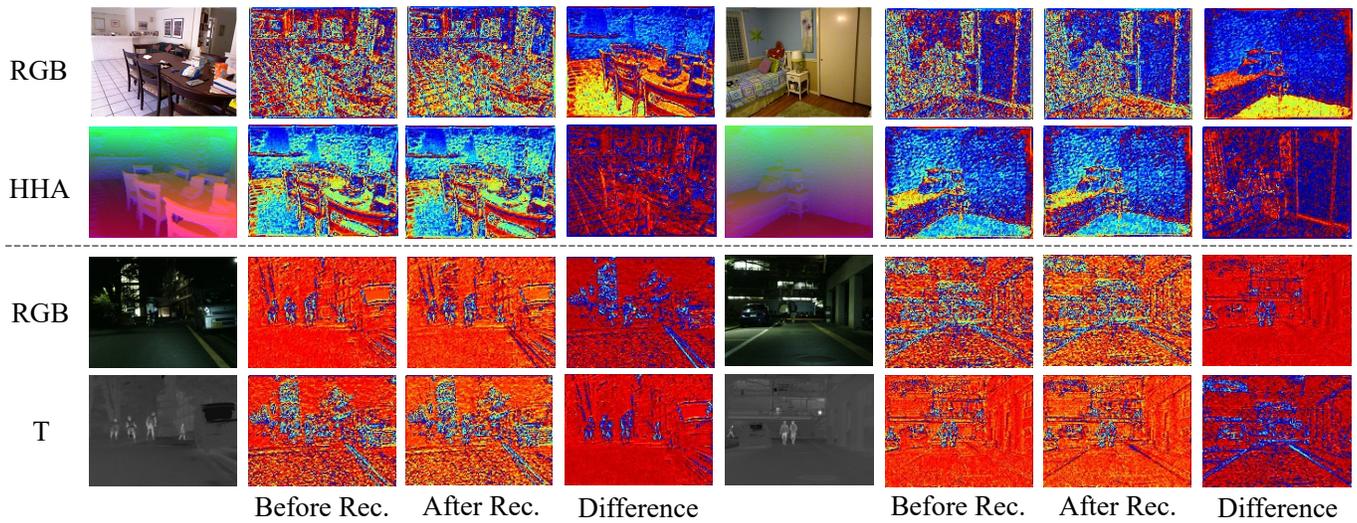}
    \caption{Visualization of the feature extracted in layer 1 and the rectified feature, and their difference map.}
    \label{fig:feature_vis}
\end{figure*}

\noindent\textbf{Feature analysis.}
To understand the key module for feature rectification, we visualize the input- and rectified features of CM-FRM in layer 1, and their difference map, as shown in Fig.~\ref{fig:feature_vis}.
It can be seen that the feature maps are enhanced in both streams after the cross-modal calibration. The RGB stream delivers texture information to the supplement modality, while the supplement modality further improves the boundary and emphasizes complementary discontinuities of RGB features.
In the RGB-D segmentation scenario, the RGB-feature difference map shows that the ground area is better spotlighted, thanks to the HHA image encoding depth information, which provides geometric cues such as height above ground, beneficial for higher-level semantic prediction of ground-related classes. 
In the RGB-T nighttime scene parsing cases, the pedestrians are hard to be seen in the RGB images. But the RGB-feature difference map clearly highlights the pedestrians thanks to the supplementary thermal modality with infrared imaging.
These indicate that the complementary features have been infused into the RGB stream. The RGB features have been rectified to better focus on informative ones and capture such complementary discontinuities towards accurate semantic understanding.

\end{document}